\documentclass[11pt]{article}

\usepackage[final]{acl}

\usepackage{times}
\usepackage{latexsym}

\usepackage[T1]{fontenc}

\usepackage[utf8]{inputenc}

\usepackage{microtype}

\usepackage{inconsolata}

\usepackage{graphicx}
\usepackage{booktabs}
\usepackage{amsfonts}
\usepackage{amsmath}
\usepackage[table]{xcolor}
\usepackage{hyperref}
\usepackage{multirow}
\usepackage[ruled,vlined]{algorithm2e}
\usepackage{subcaption}
\usepackage{pgffor} 
\usepackage{alphalph}

\usepackage{makecell}
\usepackage{natbib} 
\usepackage{setspace}
\usepackage{pifont}
\usepackage{amssymb}

\definecolor{babyblueeyes}{rgb}{0.63, 0.79, 0.95}
\definecolor{babyblue}{rgb}{0.54, 0.81, 0.94}
\definecolor{bluegray}{rgb}{0.4, 0.6, 0.8}
\definecolor{brightgreen}{rgb}{0.4, 1.0, 0.0}
\definecolor{harlequin}{rgb}{0.25, 1.0, 0.0}
\definecolor{harvardcrimson}{rgb}{0.79, 0.0, 0.09}
\newcommand{\cc}{\cellcolor{babyblueeyes!20}}

\title{Unleashing Low-Bit Inference on Ascend NPUs: A Comprehensive Evaluation of HiFloat Formats}

\author{Pengxiang Zhao, Hui-Ling Zhen, Xing Li, Han Bao, Weizhe Lin, Zhiyuan Yang, Manyi Zhang,  \\ 
\textbf{Yuanyong Luo, Ziwei Yu, Xin Wang, Mingxuan Yuan, Xianzhi Yu, Zhenhua Dong } \\
\AND
\textnormal{Huawei Technologies Co., Ltd.}
} 

\begin{document}
\maketitle
\begin{abstract}
As LLMs scale, low-bit floating-point formats like MXFP and NVFP4 offer new opportunities for precision and efficiency. In this work, we evaluate HiFloat (HiF8 and HiF4), a family of formats tailored for Ascend NPUs. Through rigorous comparison across weight-activation and KV-cache tasks, we provide three key insights: (1) INT8 suits narrow-range data, while floating-point formats excel with high-variance data; (2) in 4-bit regimes, HiF4’s hierarchical scaling prevents the accuracy collapse seen in integer formats; and (3) HiFloat is fully compatible with state-of-the-art post-training quantization frameworks. Overall, HiFloat provides a solution for high-efficiency LLM inference on NPUs.
\end{abstract}

\section{Introduction}
Exponential scaling in Large Language Models (LLMs) has placed immense pressure on computational throughput and memory bandwidth. Quantization has emerged as a critical paradigm to alleviate these bottlenecks by reducing memory overhead without significantly sacrificing performance. Current literature has largely focused on integer-based methods \citep{frantar2022gptq, lin2023awq, zhao2025ganq, xiao2023smoothquant, shaoomniquant, li2024svdquant, KIVI, hooper2025kvquant, li2025kvtuner, kim2024squeezellm, ashkboos2024quarot, maaffinequant, liu2025spinquant, zhang2024mixpe} for weight, activation, and KV cache quantization to enhance inference efficiency.

Building on these foundations, the field is increasingly pivoting toward low-bit floating-point formats. Hardware-native standards like Microscaling (MX)~\citep{rouhani2023microscalingdataformatsdeep} and NVFP4~\citep{nvidia_nvfp4} represent a significant step in this evolution. For example, GPT-OSS~\citep{openai2025gptoss120bgptoss20bmodel} leverages MXFP4 quantization-aware training (QAT) to compress weights, enabling single-GPU execution for large-scale models.

In this work, we evaluate the HiFloat format family~\citep{luo2024hifloat8, luo2026hifloat4} tailored for Ascend NPUs. The 8-bit HiFloat (HiF8) features a dynamic mantissa allocation for extended range, while the 4-bit HiFloat (HiF4) employs a three-level hierarchical scaling structure. While HiF8 has been explored for training~\citep{luo2024hifloat8}, the efficacy of the HiFloat family for post-training quantization (PTQ) remains unexamined. We address this gap with a comprehensive evaluation of their distributional properties, conversion accuracy, and end-to-end performance.

Furthermore, we explore the synergy between HiFloat and established PTQ frameworks like SmoothQuant \citep{xiao2023smoothquant} and SVDQuant \citep{li2024svdquant}. Our analysis demonstrates that these outlier-mitigation strategies complement HiFloat’s inherent representational capacity, yielding compounded benefits for model compression. Beyond weight-activation quantization, we evaluate HiFloat for KV cache quantization. The results show that HiF4 achieves state-of-the-art performance, outperforming both 4-bit integer baselines and existing floating-point standards.

Our contributions are summarized as follows:
\begin{itemize}
    \item \textbf{Mathematical Formulation:} We provide formal quantization and dequantization logic for the HiFloat family.
    \item \textbf{Distributional Analysis:} We analyze the representational properties of HiFloat formats.
    \item \textbf{Inference Evaluation:} We present the first comprehensive study of HiFloat for LLM inference, evaluating performance across the entire pipeline, including weights, activations, and KV cache.
    \item \textbf{Algorithmic Synergy Analysis:} We characterize the effective interplay between HiFloat and established PTQ frameworks like SmoothQuant and SVDQuant.
\end{itemize}
\section{Background and Related Work}
\paragraph{Floating-point Representation.}{Floating-point (FP) representation is fundamental to modern deep learning, offering the dynamic range necessary to capture high-variance data distributions. Following the IEEE 754 standard~\citep{ieee754_2019}, an FP number comprises a sign bit ($s$), an exponent ($e$), and a mantissa ($m$). We adopt the E$x$M$y$ notation to indicate that $x$ bits are allocated to the exponent and $y$ bits to the mantissa.}

\paragraph{The Microscaling Formats.}{The Microscaling (MX) specification~\citep{rouhani2023microscalingdataformatsdeep} introduces block-based quantization formats, which partitions tensors into contiguous micro-blocks that share a single, high-precision scale factor. An MX representation is formally defined by the triplet $(k, \mathcal{S}, \mathcal{E})$, where $k$ is the block size (typically 32), $\mathcal{S}$ denotes the shared scaling data type (typically E8M0), and $\mathcal{E}$ is the element data type. We detail MX types and conversion logic in Appendix~\ref{sec:to_mxfp}.}

\paragraph{The NVFP4 Format.}{The NVFP4 \citep{nvidia_nvfp4} extends the MX paradigm using a two-level hierarchical scaling mechanism. Each 16-element micro-block is scaled by a fine-grained E4M3 factor, which is then modulated by a per-tensor FP32 scalar to maintain numerical fidelity. We detail its configuration and conversion logic in Appendix \ref{sec:to_nvfp}.}

\paragraph{PTQ Methods.}{PTQ methods quantize pretrained models without extensive retraining, targeting three primary operands: weights \citep{frantar2022gptq, lin2023awq}, weight-activations \citep{xiao2023smoothquant, li2024svdquant, zhang2024mixpe}, and KV cache \citep{KIVI, li2025kvtuner}. To mitigate accuracy degradation caused by heavy-tailed distributions in LLMs, techniques such as outlier splitting~\citep{kim2024squeezellm, li2024svdquant}, magnitude smoothing~\citep{lin2023awq, wei2023outlier, xiao2023smoothquant}, and coordinate rotation~\citep{ashkboos2024quarot, maaffinequant, liu2025spinquant, shao2025blockrotation} are leveraged to suppress outliers and normalize feature distributions and preserve accuracy. While PTQ methodologies have matured for integer quantization, their efficacy remains under-explored in floating-point domains.}
\section{The 8-bit HiFloat Format: HiF8}\label{sec:hif8}
\begin{figure}[!t]
    \centering
    \includegraphics[width=0.9\linewidth]{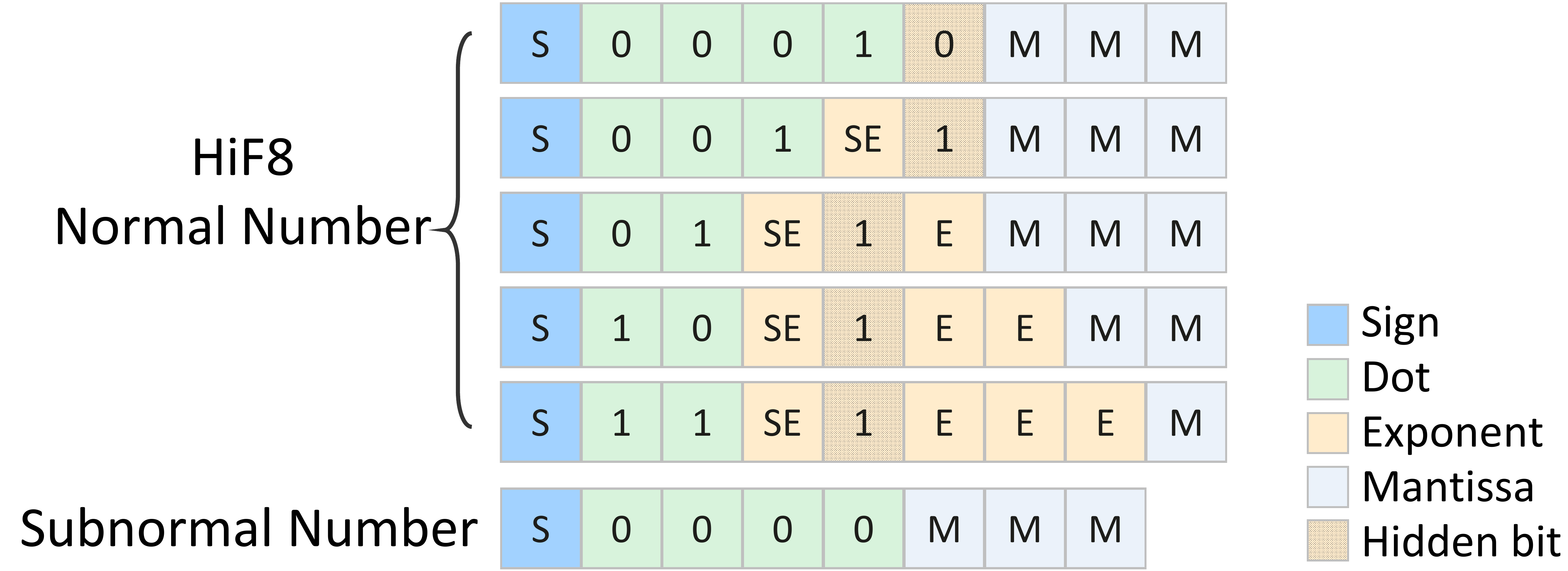}
    \caption{HiF8 Bit Layout.}
    \label{fig:hif8}
\end{figure}

The 8-bit HiFloat format (HiF8)~\citep{luo2024hifloat8} extends the IEEE 754 standard by introducing a dynamic prefix code to adapt bit-field allocation to numerical distributions. This design provides the representational flexibility required for diverse workloads through four distinct components (Figure~\ref{fig:hif8}). The sign field determines polarity, while the dot field distinguishes between normal and subnormal forms and dictates the dynamic bit-width split between the exponent and mantissa. The exponent field is context-dependent: it represents either an implicit zero or an exponent sign bit paired with an implicit leading bit of $1$. This implicit bit is restored during decoding to maximize precision. For example, the configuration $\{\text{SE}, [1]\}$ represents $\pm 1$, while $\{\text{SE}, [1], \text{E}\}$ covers $\{\pm 2, \pm 3\}$. All remaining bits are allocated to the mantissa.

\begin{table}[!t]
\centering
\renewcommand{\arraystretch}{1.0}
\setlength{\tabcolsep}{1pt}
\resizebox{0.9\linewidth}{!}{
\begin{tabular}{@{\hspace{1em}}lccc@{\hspace{1em}}}
\toprule
\textbf{Metric} & \textbf{E4M3} & \textbf{E5M2} & \textbf{HiF8} \\ \midrule
Max Precision (Bits) & 3 & 2  & 3 \\
Max Positive Normal & $1.75\times2^8$ & $1.75\times2^{15}$ & $2^{15}$ \\
Min Positive Normal & $2^{-6}$ & $2^{-14}$ & $2^{-15}$ \\ 
Max Positive Subnormal & $1.75\times2^{-7}$ & $1.5\times2^{-16}$ & $2^{-16}$ \\
Min Positive Subnormal & $2^{-9}$ & $2^{-16}$ & $2^{-22}$ \\ \bottomrule
\end{tabular}
}
\caption{Comparison across FP8 Formats.}
\label{tab:fp8_comparison}
\end{table}

\begin{figure*}[!t]
    \centering
    \pagebreak[0] 
    \vspace{1em}
    \begin{subfigure}{0.45\textwidth}
        \centering
            \includegraphics[width=\textwidth]{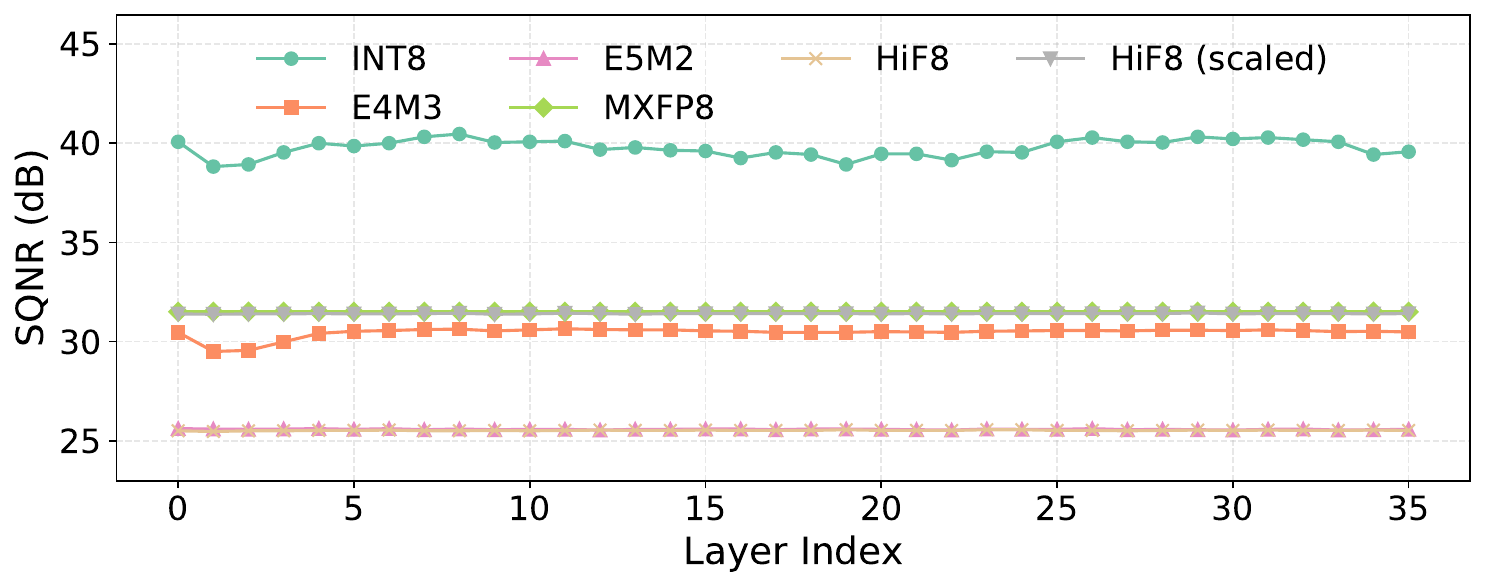}
            \caption{Weight} \label{fig:sqnr_8bit_w}
    \end{subfigure}
    \begin{subfigure}{0.45\textwidth}
        \centering
            \includegraphics[width=\textwidth]{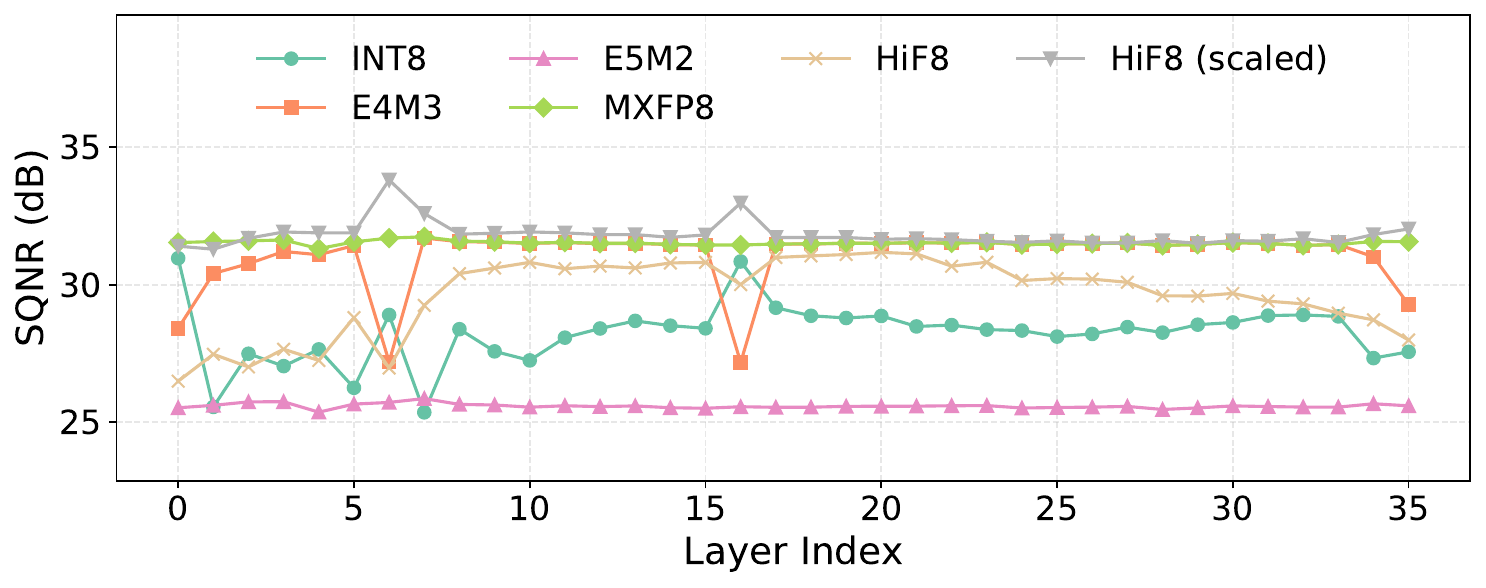}
            \caption{Activation} \label{fig:sqnr_8bit_a}
    \end{subfigure}
    \begin{subfigure}{0.45\textwidth}
        \centering
            \includegraphics[width=\textwidth]{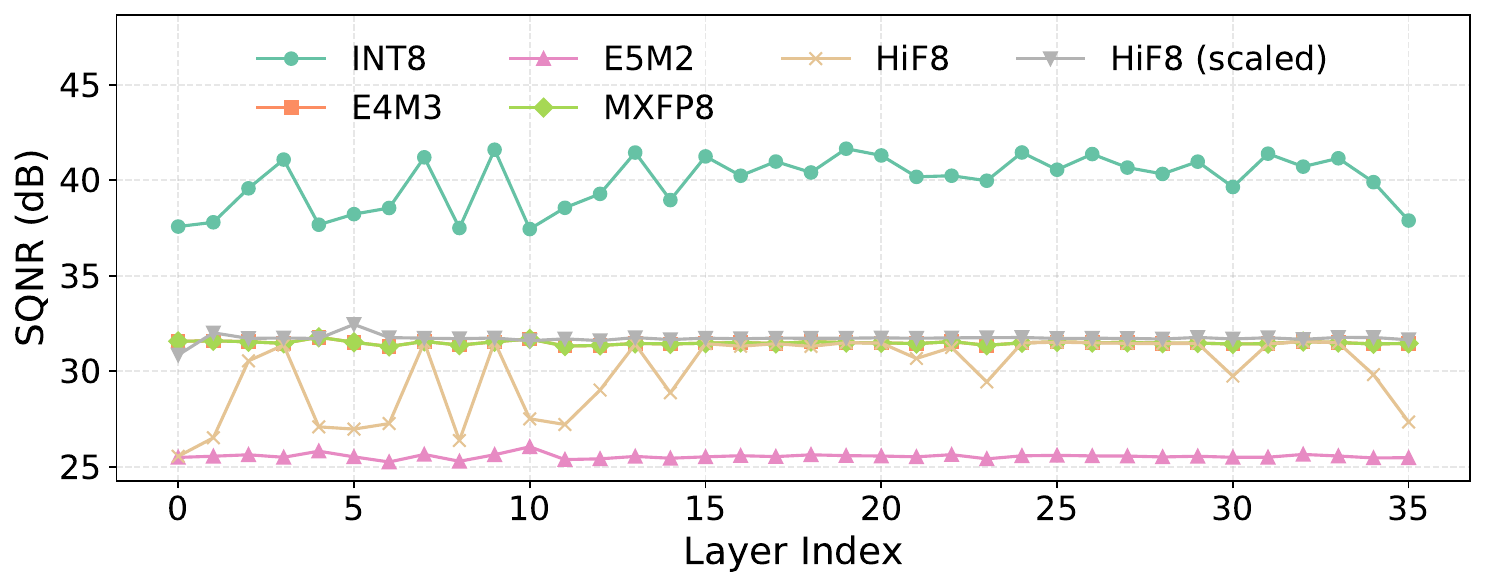}
            \caption{Key State} \label{fig:sqnr_8bit_k}
    \end{subfigure}
    \begin{subfigure}{0.45\textwidth}
        \centering
            \includegraphics[width=\textwidth]{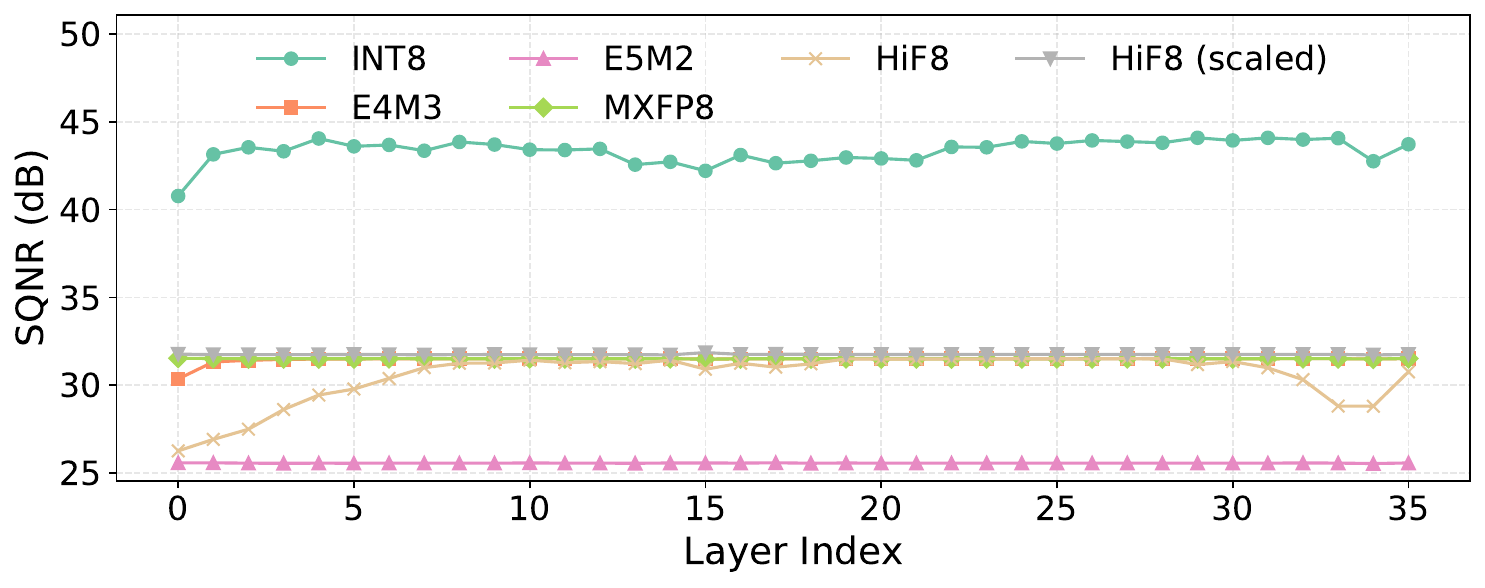}
            \caption{Value State} \label{fig:sqnr_8bit_v}
    \end{subfigure}
    \caption{Layer-wise Quantization SQNR ($\uparrow$) Comparison for Qwen3-8B across 8-bit Formats.}
    \label{fig:layer_wise_sqnr_qwen3_8b_8bit_format}
\end{figure*}

Adhering to the IEEE 754 standard, the value of a HiF8 normal number is defined by: 
\begin{equation}\label{equ:hif8_normal}
(-1)^s \times (1 + m) \times 2^{e}, 
\end{equation} 
whereas for subnormal numbers, the value is represented as: 
\begin{equation} 
 (-1)^s \times 2^{m-23}. 
\end{equation}
The dynamic range of HiF8 is summarized in Table~\ref{tab:fp8_comparison}, while the formal mapping procedure from high-precision inputs is detailed in Appendix~\ref{sec:to_hif8}. 

While HiF8 performance is established for training~\citep{luo2024hifloat8}, its fidelity for static quantization remains unexamined. We address this by evaluating how various 8-bit formats preserve signal integrity using the Signal-to-Quantization-Noise Ratio (SQNR):
\begin{equation*}
    \text{SQNR}(\mathbf{X}, \hat{\mathbf{X}}) = 10\log_{10}\left( \frac{\|\mathbf{X}\|_F^2}{\|\mathbf{X} - \hat{\mathbf{X}}\|_F^2} \right),
\end{equation*}
where $\mathbf{X}$ and $\hat{\mathbf{X}}$ are the original and quantized weight matrices, and $\| \cdot \|_F$ is the Frobenius norm. A higher SQNR signifies superior fidelity.

\begin{figure}
    \centering
    \includegraphics[width=\linewidth]{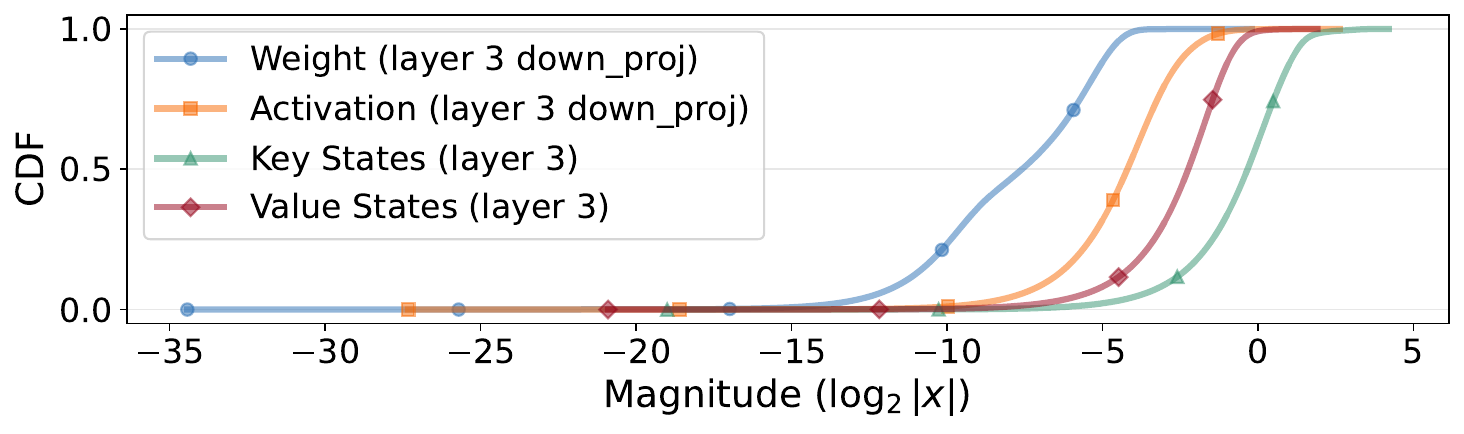}
    \caption{CDF of Qwen3-8B Representative Cases.}
    \label{fig:cdf}
\end{figure}

\begin{figure}
    \centering
    \includegraphics[width=\linewidth]{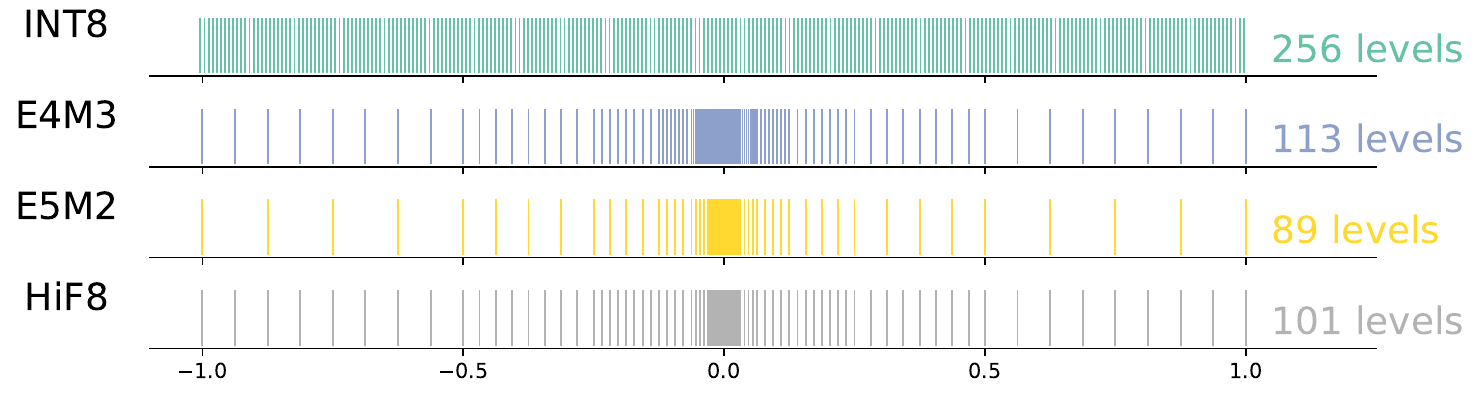}
    \caption{The Distribution of Representable Points within the interval $[-1, 1]$ of 8-bit Formats. }
    \label{fig:density}
\end{figure}

\paragraph{Weight Fidelity.}{
Figure~\ref{fig:sqnr_8bit_w} illustrates a consistent 8-bit SQNR hierarchy across Qwen3-8B~\citep{qwen3technicalreport} weights: INT8 (per-channel) $>$ HiF8 (scaled) $\approx$ MXFP8 (E4M3) $>$ E4M3 $>$ E5M2 $\approx$ HiF8. Detailed per-weight metrics are in Appendix~\ref{appendix:sqnr_weight}. This trend reveals a fundamental shift in numerical requirements from training to inference. While HiF8’s expansive dynamic range captures volatile gradients during training \citep{luo2024hifloat8}, it is suboptimal for the static, bounded distributions of pre-trained weights. As shown in Figure~\ref{fig:cdf}, weights concentrate heavily within $[2^{-9}, 2^{-5}]$, with negligible density in the extreme ranges where HiF8 allocates significant representational capacity. This mismatch results in bit waste. Within the primary range ($[-1, 1]$), INT8 utilizes its full 8-bit budget (256 levels), whereas E4M3, HiF8, and E5M2 provide only 113, 101, and 89 levels, respectively (Figure~\ref{fig:density}). Consequently, HiF8 exhibits increasing discretization coarseness as magnitudes deviate from zero (see Appendix~\ref{sec:dist_analysis_8bit}), sacrificing the fine-grained precision necessary for high-fidelity weight compression. To mitigate this, a per-channel scaling factor could be employed to realign the target tensor's range with HiF8’s capacity; the specific algorithm for this approach is detailed in Appendix~\ref{sec:scale_hif8}. Denoted as HiF8 (scaled), such a configuration ($K=16$) yield performance comparable to that of MXFP8. A comparison of scaling granularity and bit-width is provided in Table~\ref{tab:8_bit_format_comparison}. MXFP8 typically requires a higher bit-budget per element compared to the HiF8 variants.
}

\begin{table}[!t]
\centering
\small
\setlength{\tabcolsep}{6pt}
\resizebox{0.95\linewidth}{!}{
\begin{tabular}{l cccc}
\toprule
\textbf{Feature} & \textbf{INT8} & \textbf{MXFP8} & \textbf{HiF8} & \textbf{HiF8 (scaled)} \\ \midrule
Scaling Granularity & Per-axis & Block-based & None & Per-axis \\
Block Size & $N$ & 32 & --- & $N$ \\
\midrule
Scale Format & BF16 & E8M0 & --- & FP32 \\
Element Format & INT8 & E4M3 & HiF8 & HiF8 \\
\midrule
Bits/Element & $\gtrsim 8.00$ & $8.25$ & $8.00$ & $\gtrsim 8.00$ \\ \bottomrule
\end{tabular}
}
\caption{Comparison of 8-bit Quantization Formats.}
\label{tab:8_bit_format_comparison}
\end{table}

\paragraph{Activation Fidelity.}{Unlike weights, activations exhibit high dynamic volatility and emergent outliers. We evaluate 8-bit activation fidelity using Qwen3-8B on Wikitext-2 dataset~\citep{merity2016pointer}. As shown in Figure~\ref{fig:sqnr_8bit_a}, there is HiF8 (scaled) $\approx$ MXFP8 (E4M3) $>$ E4M3 $>$ HiF8 $>$ INT8 (per-token) $>$ E5M2. Detailed per-weight's activation comparisons are provided in Appendix~\ref{appendix:sqnr_activation}. Symmetric floating-point formats underutilize representable states due to activation asymmetry, wasting bit-budget on empty numerical regions; however, their wide dynamic range inherently accommodates extreme outliers. Conversely, while INT8 captures distributional shifts via zero-points, large-magnitude outliers force a coarse, uniform quantization step that significantly degrades resolution near zero, where activation density is highest. With $K=4$, HiF8 (scaled) achieves performance comparable to that of MXFP8. A detailed distributional analysis is provided in Appendix~\ref{sec:dist_analysis_8bit}.

Ultimately, end-to-end performance depends on the complex interplay between weight and activation precision. For instance, the superior fidelity of INT8 for weights may be offset by its inability to capture activation dynamics. We evaluate the end-to-end quantization performance in Section~\ref{sec:exp}.
}

\paragraph{KV Cache Fidelity.}{We evaluate KV cache fidelity by applying per-token quantization to Key and Value tensors across 64 Wikitext-2 sequences. As shown in Figures~\ref{fig:sqnr_8bit_k} and~\ref{fig:sqnr_8bit_v}, both tensors follow a consistent SQNR hierarchy: INT8 (per-token) $>$ HiF8 (scaled) $\approx$ MXFP8 (E4M3) $>$ E4M3 $>$ HiF8 $>$ E5M2.

This hierarchy suggests that the linear projections generating Key and Value states effectively mitigate the emergent outliers observed in raw activations (Figures~\ref{fig:dist_8bit_k} and~\ref{fig:dist_8bit_v} in Appendix~\ref{sec:dist_analysis_8bit}). Thus, the uniform resolution of INT8 is better suited for these narrower ranges than the expansive, but coarser, dynamic range of HiF8 or E5M2. With $K=1$, HiF8 (scaled) achieves performance comparable to that of MXFP8.
}

\section{The 4-bit HiFloat Format: HiF4}\label{sec:hif4}
\begin{figure}[!t]
    \centering
    \includegraphics[width=0.9\linewidth]{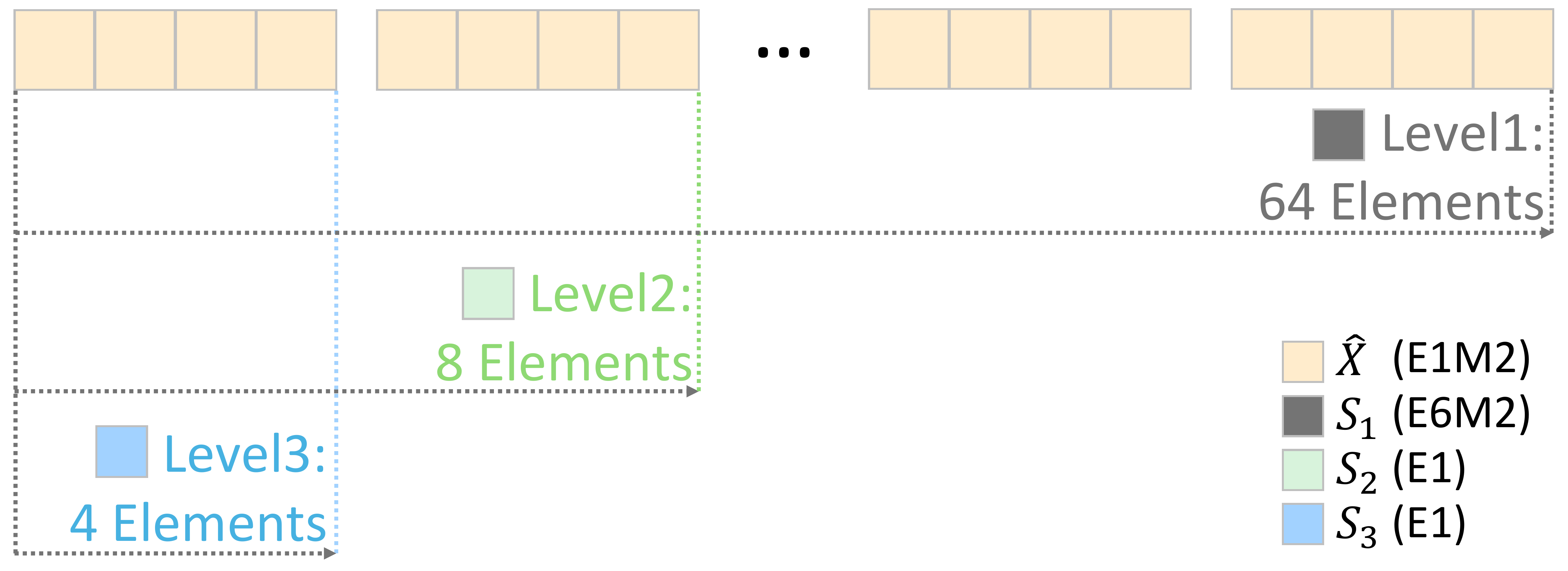}
    \caption{HiF4 Hierarchical Quantization.}
    \label{fig:hif4}
\end{figure}

\begin{table}[!t]
\centering
\small
\setlength{\tabcolsep}{6pt}
\resizebox{0.9\linewidth}{!}{
\begin{tabular}{l ccc}
\toprule
\textbf{Feature} & \textbf{MXFP4} & \textbf{NVFP4} & \textbf{HiF4} \\ \midrule
Hierarchy Levels & 1 & 2 & 3 \\
Per-tensor Scaling & \ding{55} & \ding{51} & \ding{55}\\
Block Size & 32 & 16 & 64 \\
Sub-lock Size & \ding{55} & \ding{55} & 8 \\
Micro-block Size & \ding{55} & \ding{55} & 4 \\\midrule
Scaling Format & E8M0 & FP32/E4M3 & E6M2/E1/E1 \\
Element Format & E2M1 & E2M1 & E1M2 \\
\midrule
Bits/Element& 4.25 & $\gtrsim$ 4.50 & 4.50 \\ \bottomrule
\end{tabular}
}
\caption{Comparison of 4-bit Hierarchical Quantization.}
\label{tab:format_comparison}
\end{table}

\begin{figure*}
    \centering
    \pagebreak[0]
    \vspace{1em}
    \begin{subfigure}{0.45\textwidth}
        \centering
            \includegraphics[width=\textwidth]{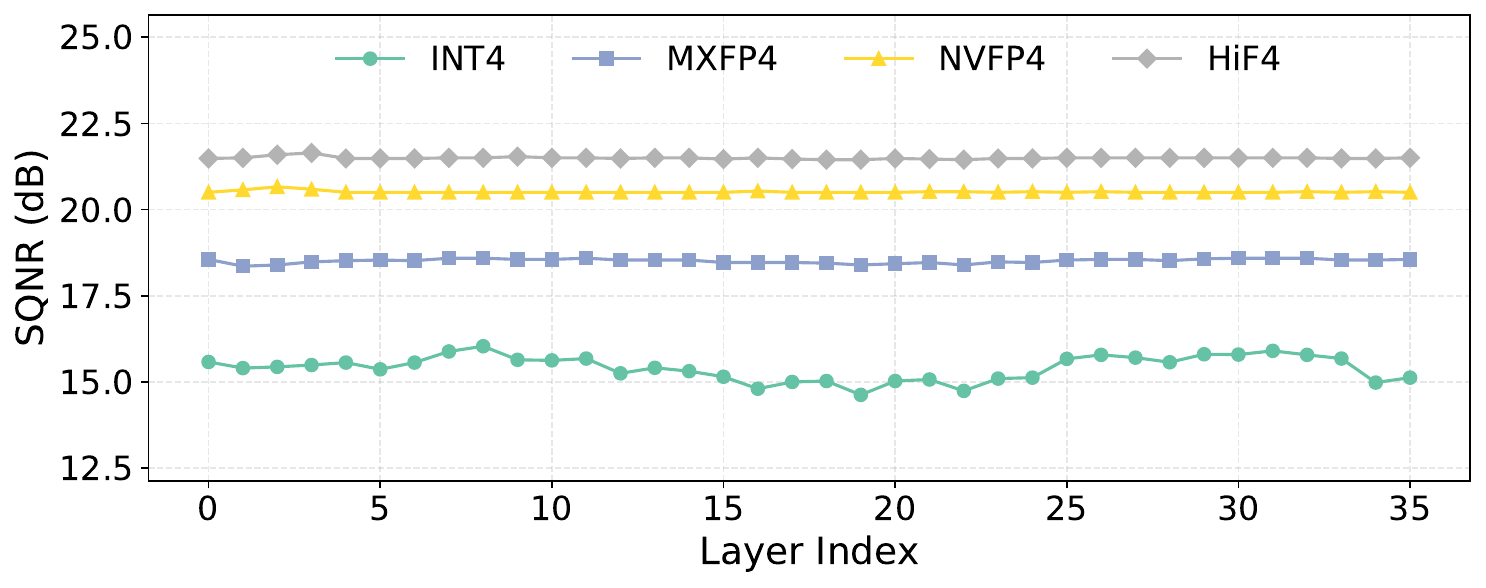}
            \caption{Weight} \label{fig:sqnr_4bit_w}
    \end{subfigure}
    \begin{subfigure}{0.45\textwidth}
        \centering
            \includegraphics[width=\textwidth]{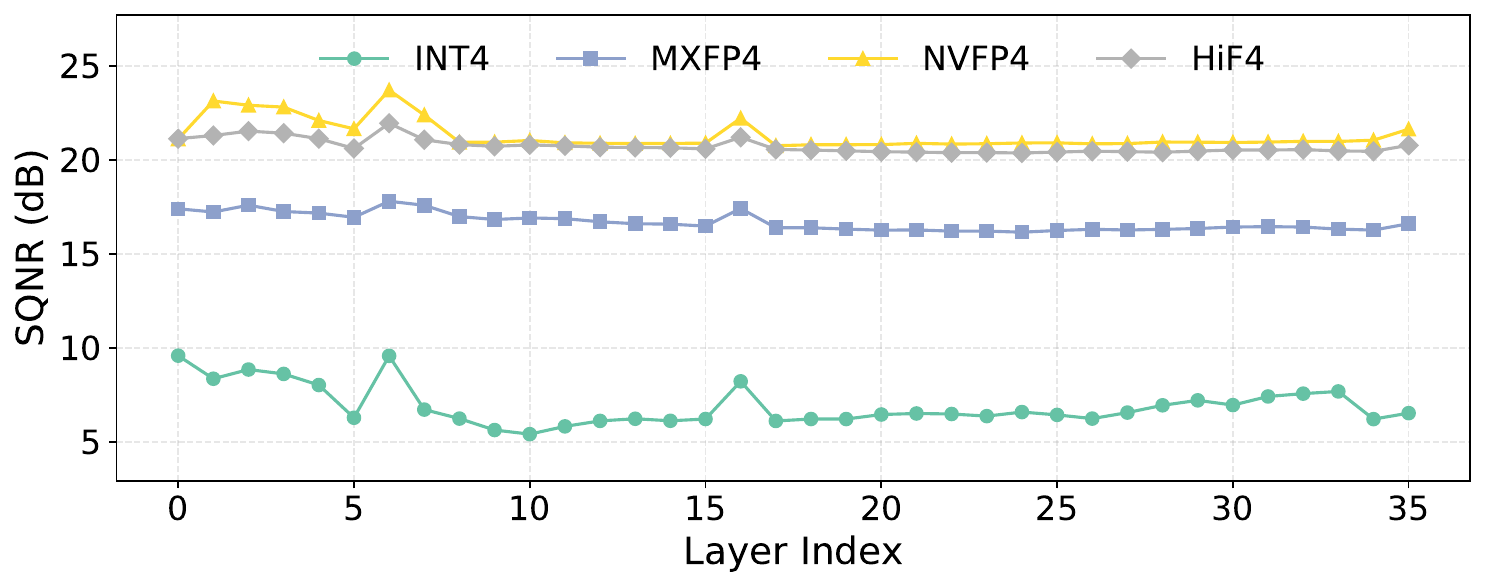}
            \caption{Activation} \label{fig:sqnr_4bit_a}
    \end{subfigure}
    \begin{subfigure}{0.45\textwidth}
        \centering
            \includegraphics[width=\textwidth]{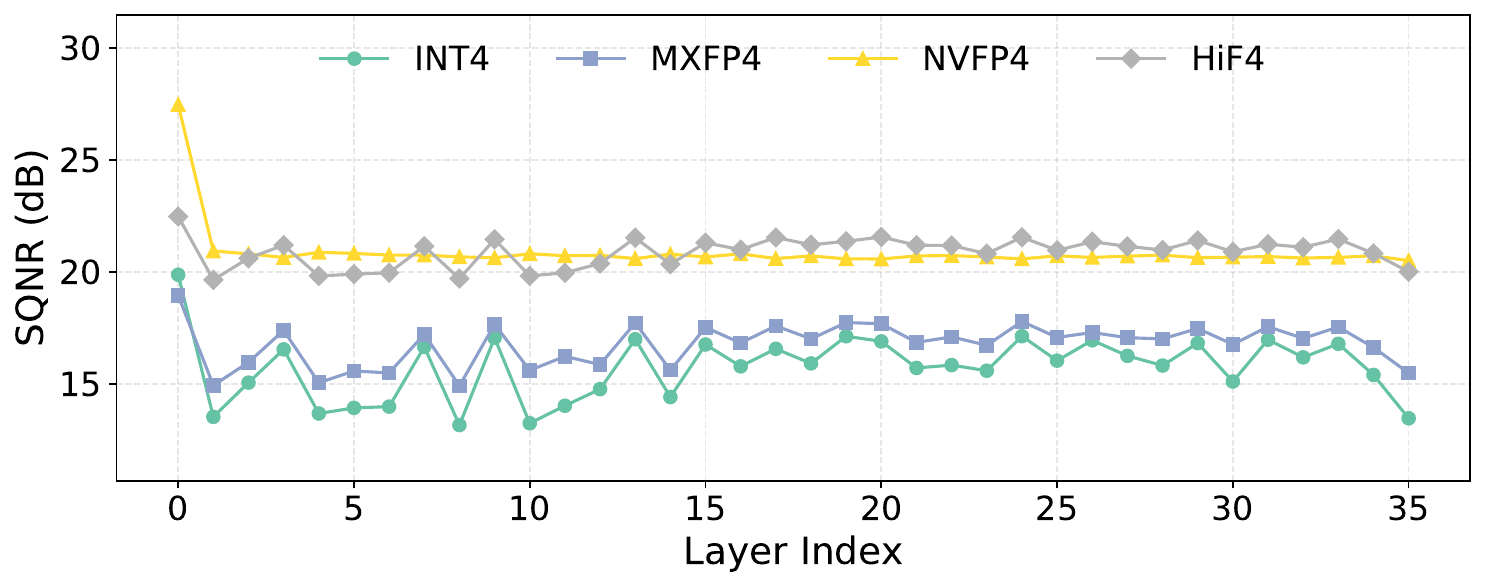}
            \caption{Key State}\label{fig:sqnr_4bit_k}
    \end{subfigure}
    \begin{subfigure}{0.45\textwidth}
        \centering
            \includegraphics[width=\textwidth]{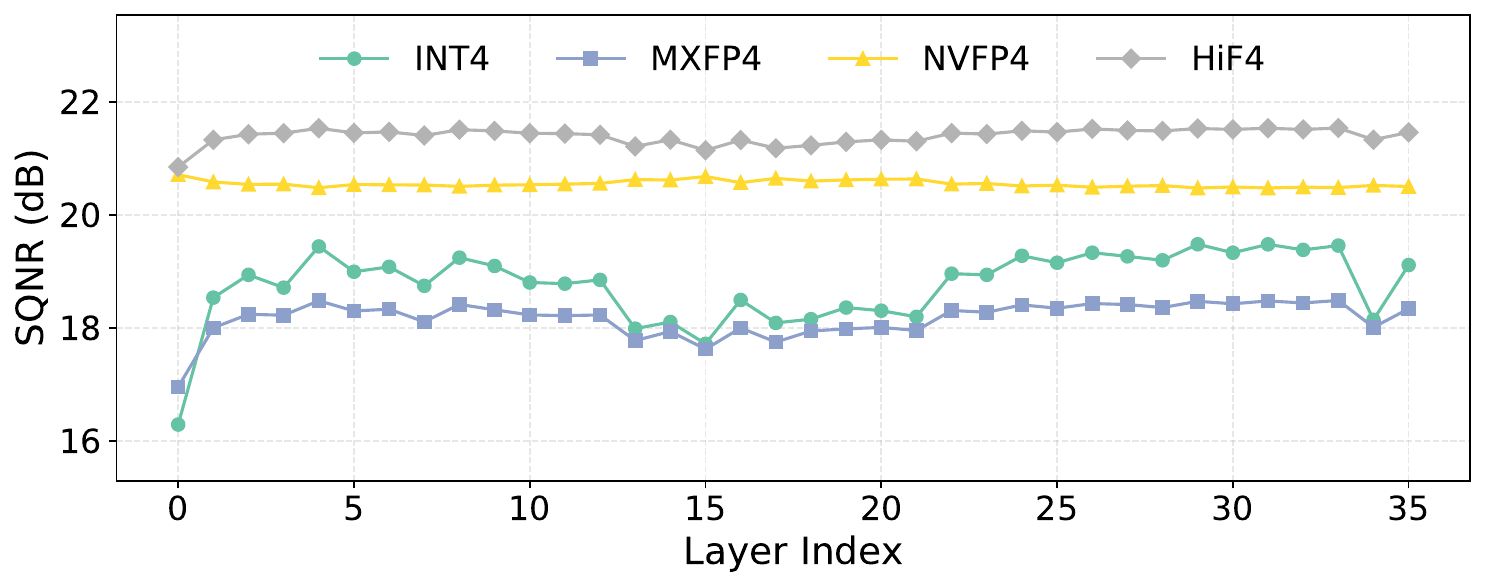}
            \caption{Value State}\label{fig:sqnr_4bit_v}
    \end{subfigure}
    \caption{Layer-wise Quantization SQNR ($\uparrow$) Comparison for Qwen3-8B across 4-bit Formats.}
    \label{fig:layer_wise_sqnr_qwen3_8b_4bit_format}
\end{figure*}

As shown in Figure~\ref{fig:hif4}, The 4-bit HiFloat format (HiF4)~\citep{luo2026hifloat4} employs a three-level hierarchical encoding scheme. Each 64-element block shares an 8-bit unsigned E6M2 scale, which is refined by 1-bit E1M0 factors at both the 8-element sub-block and 4-element micro-block levels. The individual elements are then represented in a 4-bit E1M2 format. A summary of the 4-bit formats is provided in Table~\ref{tab:format_comparison}. The formal procedure for mapping high-precision inputs to the HiF4 representation is detailed in Appendix~\ref{sec:to_hif4}. 

\paragraph{Weight Fidelity.}{Halving the bit-budget to 4 bits creates a critical representational bottleneck. As shown in Figure~\ref{fig:sqnr_4bit_w}, a consistent performance hierarchy emerges across Qwen3-8B layers: HiF4 $>$ NVFP4 $>$ MXFP4 $>$ INT4 (per-channel). Detailed per-weight metrics are provided in Appendix~\ref{appendix:sqnr_weight}. This hierarchy demonstrates that at ultra-low precisions, hierarchical block-based scaling is essential to capture local variance and recover representational capacity. While INT4 suffers from extreme binning coarseness, the multi-level scaling of HiF4 and NVFP4 provides a significantly finer discretization grid than the single-level microscaling of MXFP4. A qualitative distributional analysis is provided in Appendix~\ref{sec:dist_analysis_4bit}.
}

\paragraph{Activation Fidelity.}{As shown in Figure~\ref{fig:sqnr_4bit_a}, the 4-bit activation SQNR hierarchy follows: NVFP4 $>$ HiF4 $>$ MXFP4 $>$ INT4 (per-token). Detailed metrics are provided in Appendix~\ref{appendix:sqnr_activation}. The superior performance of NVFP4 is primarily driven by its incorporation of an FP32 per-tensor global scaling factor. This mechanism provides the expansive dynamic range necessary to accommodate high-magnitude activation outliers. We provide a detailed analysis of these representational trade-offs in Appendix~\ref{sec:dist_analysis_4bit}}

\paragraph{KV Cache Fidelity.}{Figures~\ref{fig:sqnr_4bit_k} and~\ref{fig:sqnr_4bit_v} illustrate 4-bit KV cache fidelity across Qwen3-8B. For Key states, NVFP4 excels in earlier layers, while HiF4 generally dominates as depth increases; in both regimes, hierarchical floating-point formats significantly outperform the uniform INT4 baseline.For Value states, the performance hierarchy shifts to: HiF4 $>$ NVFP4 $>$ INT4 (per-channel) $>$ MXFP4. This shift suggests that Value distributions are more stable and bounded than Key states; in such contexts, the uniform resolution of INT4 provides a slight advantage over the coarser, single-level scaling of MXFP4. These observations align with the representable distributions analyzed in Appendix~\ref{sec:dist_analysis_4bit}, where the finer granularity of HiF4 and NVFP4 consistently mitigates the coarse binning seen in INT4 and MXFP4.
}
\section{Experimental Results}\label{sec:exp}
We evaluate the end-to-end performance of HiFloat formats across multiple models and assess their synergy with existing PTQ frameworks.

\subsection{Settings}
Following prior work, we evaluate model performance using perplexity on Wikitext-2~\citep{merity2016pointer} and C4~\citep{raffel2023exploring} with a 2,048 sequence length.
Zero-shot accuracy is further assessed across adiverse suite, including ARC Challenge~\citep{clark2018thinksolvedquestionanswering}, HellaSwag~\citep{zellers2019hellaswagmachinereallyfinish}, MMLU~\citep{hendrycks2021measuring}, Math-500~\citep{hendrycks2021measuring}, GSM8K~\citep{cobbe2021trainingverifierssolvemath}, and LongBench~\citep{bai2023longbench}, utilizing the LM Evaluation Harness library~\citep{eval-harness}. For SmoothQuant, we optimize the hyperparameter $\alpha$ via grid search across $[0.1, 0.9]$ with a $0.1$ step size. For SVDQuant, we fix the low-rank component at rank $16$.

\begin{table*}[!t]
\centering
\small
\renewcommand{\arraystretch}{1.1}
\setlength{\tabcolsep}{6pt}
\resizebox{0.9\linewidth}{!}{
\begin{tabular}{c|c|c|c|cc|cccc|cc}
\toprule
\multirow{2}{*}{\textbf{Models}} & \multirow{2}{*}{\textbf{Bit}} & \multirow{2}{*}{\textbf{Methods}} & \multirow{2}{*}{\textbf{Format}} & \multicolumn{2}{c|}{\textbf{PPL} ($\downarrow$)} & \multicolumn{4}{c|}{\textbf{Acc} ($\uparrow$)} & \multicolumn{2}{c}{\textbf{Overall}} \\
\cline{5-12}
& & & & \textbf{Wiki} & \textbf{C4} & \textbf{HellaSwag} & \textbf{MMLU} & \textbf{Arc-C} & \textbf{MATH-500} & \textbf{Avg.}$\uparrow$ & $\Delta$(\%)$\downarrow$ \\
\midrule
\multirow{20}{*}{\textbf{Qwen3-8B}} 
& - & - & BF16 & 9.72 & 15.43 & 0.571 & 0.730 & 0.555 & 0.784 & 1.000 & - \\ 
\cline{2-12}

& \multirow{9}{*}{W8A8} & \multirow{3}{*}{RTN} 
& INT & \textbf{9.63} & \textbf{15.40} & \textbf{0.572} & \textbf{0.727} & 0.547 & \textbf{0.800} & \textbf{1.002} & \textbf{-0.2} \\
& & & MXFP & 9.76 & 15.48 & 0.570 & \textbf{0.727} & \textbf{0.549} & 0.796 & 1.000 & 0.0 \\
& & & \cc HiF & \cc 9.67 & \cc 15.41 & \cc 0.569 & \cc 0.723 & \cc 0.547 & \cc 0.792 & \cc 0.997 & \cc 0.3 \\
\cline{3-12}

& & \multirow{3}{*}{SmoothQuant} 
& INT & \textbf{9.59} & \textbf{15.29} & \textbf{0.572} & \textbf{0.728} & 0.548 & 0.792 & 1.000 & 0.0 \\
& & & MXFP & 9.75 & 15.48 & 0.570 & \textbf{0.728} & \textbf{0.561} & 0.786 & \textbf{1.001} & \textbf{-0.1} \\
& & & \cc HiF & \cc 9.79 & \cc 15.47 & \cc 0.568 & \cc 0.723 & \cc 0.554 & \cc \textbf{0.798} & \cc \textbf{1.001} & \cc \textbf{-0.1} \\
\cline{3-12}

& & \multirow{3}{*}{SVDQuant} 
& INT & \textbf{9.58} & \textbf{15.28} & \textbf{0.573} & \textbf{0.729} & \textbf{0.552} & \textbf{0.796} & \textbf{1.003} & \textbf{-0.3} \\
& & & MXFP & 9.70 & 15.40 & 0.571 & \textbf{0.729} & 0.551 & 0.792 & 1.001 & -0.1 \\
& & & \cc HiF & \cc 9.64 & \cc 15.36 & \cc 0.570 & \cc 0.728 & \cc 0.550 & \cc 0.794 & \cc 1.001 & \cc -0.1 \\

\cline{2-12}
& \multirow{12}{*}{W4A4} & \multirow{4}{*}{RTN} 
& INT & 3.6e4 & 3.7e4 & 0.260 & 0.248 & 0.196 & 0.026 & 0.277 & 72.3 \\
& & & MXFP & 11.21 & 18.13 & 0.519 & 0.652 & 0.480 & 0.720 & 0.898 & 10.2 \\
& & & NVFP & \textbf{10.16} & \textbf{16.24} & 0.551 & \textbf{0.706} & 0.521 & \textbf{0.786} & \textbf{0.971} & \textbf{2.9} \\
& & & \cc HiF & \cc 10.30 & \cc 16.55 & \cc \textbf{0.555} & \cc 0.705 & \cc \textbf{0.529} & \cc 0.760 & \cc 0.965 & \cc 3.5 \\
\cline{3-12}

& & \multirow{4}{*}{SmoothQuant} 
& INT & 6.1e2 & 6.0e2 & 0.265 & 0.234 & 0.190 & 0.014 & 0.267 & 73.3 \\
& & & MXFP & 10.90 & 17.46 & 0.535 & 0.670 & 0.492 & 0.748 & 0.926 & 7.4 \\
& & & NVFP & \textbf{10.12} & \textbf{16.21} & 0.549 & \textbf{0.708} & 0.532 & \textbf{0.806} & \textbf{0.983} & \textbf{1.7} \\
& & & HiF & \cc 10.27 & \cc 16.31 & \cc \textbf{0.557} & \cc 0.705 & \cc \textbf{0.548} & \cc 0.772 & \cc 0.978 & \cc 2.2 \\
\cline{3-12}

& & \multirow{4}{*}{SVDQuant} 
& INT & 3.8e2 & 4.2e2 & 0.295 & 0.268 & 0.215 & 0.048 & 0.313 & 68.7 \\
& & & MXFP & 10.58 & 17.02 & 0.539 & 0.685 & 0.503 & 0.742 & 0.935 & 6.5 \\
& & & NVFP & \textbf{9.92} & \textbf{15.88} & 0.560 & \textbf{0.719} & 0.535 & \textbf{0.774} & \textbf{0.980} & \textbf{2.0} \\
& & & \cc HiF & \cc 9.98 & \cc 16.05 & \cc \textbf{0.563} &\cc  0.717 & \cc \textbf{0.539} & \cc 0.768 & \cc \textbf{0.980} & \cc \textbf{2.0} \\
\bottomrule
\end{tabular}
}
\caption{Performance Evaluation of Weight-Activation Quantization on Qwen3-8B. \textbf{Bold} indicates best performance per configuration; $\Delta$ shows degradation from BF16 baseline. }
\label{tab:wa-results-qwen}
\end{table*}

\begin{table*}[!t]
\centering
\small
\renewcommand{\arraystretch}{1.1}
\setlength{\tabcolsep}{6pt}
\resizebox{0.9\linewidth}{!}{
\begin{tabular}{c|c|c|c|cc|cccc|cc}
\toprule
\multirow{2}{*}{\textbf{Models}} & \multirow{2}{*}{\textbf{Bit}} & \multirow{2}{*}{\textbf{Methods}} & \multirow{2}{*}{\textbf{Format}} & \multicolumn{2}{c|}{\textbf{PPL} ($\downarrow$)} & \multicolumn{4}{c|}{\textbf{Acc} ($\uparrow$)} & \multicolumn{2}{c}{\textbf{Overall}} \\
\cline{5-12}
& & & & \textbf{Wiki} & \textbf{C4} & \textbf{HellaSwag} & \textbf{MMLU} & \textbf{Arc-C} & \textbf{MATH-500} & \textbf{Avg.}$\uparrow$ & $\Delta$(\%)$\downarrow$ \\
\midrule
\multirow{20}{*}{\textbf{openPangu-7B}} 
& - & - & BF16 & 34.95 & 57.18 & 0.450 & 0.567 & 0.323 & 0.898 & 1.000 & - \\ 
\cline{2-12}

& \multirow{9}{*}{W8A8} & \multirow{3}{*}{RTN} 
& INT & 38.76 & 63.02 & \textbf{0.456} & 0.549 & \textbf{0.331} & 0.890 & \textbf{0.995} & \textbf{0.5} \\
& & & MXFP & \textbf{34.89} & \textbf{55.67} & 0.446 & 0.535 & 0.321 & \textbf{0.906} & 0.986 & 1.4 \\
& & & \cc HiF & \cc 35.76 & \cc 58.37 & \cc 0.447 & \cc 0.523 & \cc 0.327 & \cc 0.892 & \cc 0.978 & \cc 2.2 \\
\cline{3-12}

& & \multirow{3}{*}{SmoothQuant} 
& INT & \textbf{35.79} & \textbf{56.90} & 0.450 & \textbf{0.588} & 0.321 & 0.866 & 0.995 & 0.5 \\
& & & MXFP & 35.01 & 56.90 & 0.444 & 0.580 & 0.332 & 0.884 & 1.001 & -0.1 \\
& & & \cc HiF & \cc 36.67 & \cc 59.38 & \cc 0.448 & \cc 0.578 & \cc \textbf{0.340} & \cc \textbf{0.902} & \cc \textbf{1.014} & \cc \textbf{-1.4} \\
\cline{3-12}

& & \multirow{3}{*}{SVDQuant} 
& INT & 36.85 & 59.42 & \textbf{0.454} & 0.557 & \textbf{0.328} & 0.894 & \textbf{0.998} & \textbf{0.2} \\
& & & MXFP & \textbf{34.92} & \textbf{56.25} & 0.449 & 0.551 & 0.322 & \textbf{0.900} & 0.993 & 0.7 \\
& & & \cc HiF & \cc 35.28 & \cc 57.45 & \cc 0.449 & \cc 0.543 & \cc 0.324 & \cc 0.896 & \cc 0.989 & \cc 1.1 \\

\cline{2-12}
& \multirow{12}{*}{W4A4} & \multirow{4}{*}{RTN} 
& INT & 2.2e4 & 1.6e4 & 0.260 & 0.234 & 0.220 & 0.014 & 0.326 & 67.4 \\
& & & MXFP & 54.58 & 63.41 & 0.429 & 0.490 & 0.308 & 0.818 & 0.914 & 8.6 \\
& & & NVFP & \textbf{38.96} & \textbf{57.51} & \textbf{0.435} & 0.488 & \textbf{0.322} & \textbf{0.866} & 0.943 & 5.7 \\
& & & \cc HiF & \cc 41.32 & \cc 59.69 & \cc \textbf{0.435} & \cc \textbf{0.563} & \cc 0.310 & \cc 0.862 & \cc \textbf{0.970} & \cc \textbf{3.0} \\
\cline{3-12}

& & \multirow{4}{*}{SmoothQuant} 
& INT & 6.4e2 & 7.3e2 & 0.312 & 0.275 & 0.220 & 0.072 & 0.393 & 60.7 \\
& & & MXFP & 45.55 & 62.67 & 0.430 & 0.524 & \textbf{0.315} & 0.830 & 0.938 & 6.2 \\
& & & NVFP & \textbf{38.68} & \textbf{58.18} & 0.437 & 0.486 & 0.313 & \textbf{0.894} & 0.952 & 4.8 \\
& & & \cc HiF & \cc 39.69 & \cc 60.27 & \cc \textbf{0.443} & \cc \textbf{0.543} & \cc 0.306 & \cc 0.886 & \cc \textbf{0.973} & \cc \textbf{2.7} \\
\cline{3-12}

& & \multirow{4}{*}{SVDQuant} 
& INT & 856.2 & 724.5 & 0.313 & 0.325 & 0.249 & 0.256 & 0.510 & 49.0 \\
& & & MXFP & 46.25 & 60.85 & 0.433 & 0.513 & 0.313 & 0.824 & 0.930 & 7.0 \\
& & & NVFP & \textbf{37.52} & \textbf{57.28} & 0.439 & 0.525 & \textbf{0.321} & \textbf{0.870} & 0.962 & 3.8 \\
& & & \cc HiF & \cc 38.85 & \cc 58.42 & \cc \textbf{0.440} & \cc \textbf{0.559} & \cc 0.315 & \cc 0.864 & \cc \textbf{0.973} & \cc \textbf{2.7} \\
\bottomrule
\end{tabular}
}
\caption{Performance Evaluation of Weight-Activation Quantization on openPangu-7B. \textbf{Bold} indicates best performance per configuration; $\Delta$ shows degradation from BF16 baseline. }
\label{tab:wa-results-pangu}
\end{table*}

\subsection{Main Results}
\paragraph{W8A8 Quantization.}{Tables~\ref{tab:wa-results-qwen} and \ref{tab:wa-results-pangu} demonstrate high model resilience in the W8A8 regime. For Qwen3-8B, all formats maintain over $99\%$ of the BF16 baseline even with basic round-to-nearest (RTN), with INT8 achieving peak performance when paired with SVDQuant. While HiF8 slightly trails INT8 and MXFP8 under RTN, it achieves negligible accuracy loss when integrated with SmoothQuant or SVDQuant. Similarly, for openPangu-7B, INT8 leads under RTN quantization. SmoothQuant enables HiF8 to emerge as the top-performing format, effectively matching the BF16 baseline.}

\paragraph{W4A4 Quantization.}{Tables~\ref{tab:wa-results-qwen} and \ref{tab:wa-results-pangu} reveal a catastrophic failure for INT4 at W4A4, even when augmented with SmoothQuant and SVDQuant. This underscores that at ultra-low bit widths, uniform integer spacing becomes a fundamental bottleneck. While MXFP4 outperforms the integer baseline, it still incurs notable accuracy drops. In contrast, HiF4 and NVFP4 maintain model integrity. On Qwen3-8B, both formats remain highly competitive, with NVFP4 holding a marginal lead. For openPangu-7B, HiF4 achieves remarkable fidelity, limiting accuracy degradation to just 3.0\% under RTN and narrowing to 2.7\% with SmoothQuant or SVDQuant. This near-complete recovery of the BF16 baseline confirms that HiF4’s hierarchical structure is uniquely suited for outlier-aware optimization under stringent 4-bit constraints.}

\paragraph{KV Cache Quantization.}{Building on the success of SmoothQuant in W8A8 and W4A4 regimes, we extend our evaluation to include KV cache quantization (Table~\ref{tab:results-main_qkv_attention_quant}, Appendix~\ref{sec:additional}). At 8-bit level, Qwen3-8B remains largely unaffected by these additional steps across all formats. However, for openPangu-7B, the performance gap widens, where HiF8 exhibits a more pronounced accuracy decline than MXFP8. The transition to W4A4 augmented by Q16KV4 and QKV4 serves as a stress test. MXFP4 incurs a $10\%$ accuracy loss on Qwen3-8B and suffers a total collapse on openPangu-7B, with degradation surging beyond $60\%$. In contrast, NVFP4 and HiF4 demonstrate remarkable resilience. In particular, HiF4 consistently yields superior results for Qwen3-8B, limiting accuracy losses to $3.15\%$ for Q16KV4 and $3.92\%$ for QKV4. This trend persists for openPangu-7B, where HiF4 restricts losses to $11.03\%$ and $13.84\%$, respectively. Table~\ref{tab:results-main_qkv_attention_quant_longbench} in Appendix~\ref{sec:additional} reflects similar findings on the LongBench benchmark. These results validate HiF4 as a uniquely robust format for end-to-end inference at ultra-low bit widths.
}
\section{Conclusions} 

In this work, we present a systematic evaluation of the HiFloat formats for LLM inference, benchmarking them against industry standards such as MXFP and NVFP4. Our assessment across weights, activations, and KV cache reveals that numerical representation is the primary determinant of model resilience. We identify a clear performance bifurcation in the 8-bit regime: while floating-point formats like MXFP8 and HiF8 provide the dynamic range necessary for fluctuating activations, the uniform density of INT8 remains superior for weight quantization. This suggests that for weights, avoiding the exponent-waste of floating-point formats is more beneficial than having an expansive dynamic range. However, as precision shifts to the 4-bit regime, the limitations of uniform spacing become a fundamental bottleneck, leading to a catastrophic collapse in integer formats. In this constrained space, the hierarchical structures of NVFP4 and HiF4 prove essential for isolating outliers while maintaining local precision. Notably, HiF4 preserves over $96.5\%$ of the BF16 baseline on Qwen3-8B and $97.0\%$ on openPangu-7B using only RTN for W4A4; these results improve further when integrated with SmoothQuant or SVDQuant. When augmented with KV cache quantization, HiF4 consistently achieves superior results across all evaluated 4-bit formats, establishing it as a robust standard for low precision inference.

\bibliography{custom}

\appendix

\clearpage
\section{MXFP Transformation Logic}\label{sec:to_mxfp}
The MX formats support a diverse family of configurations by varying the bit-width and numerical interpretation of the element data type $\mathcal{E}$, as summarized in Table~\ref{tab:mx-configs}.

Mapping a high-precision tensor to the MX domain involves a two-stage transformation comprising scale estimation and element-wise quantization. In standard configurations, for a given micro-block $\mathbf{x} \in \mathbb{R}^k$,the shared scale $s=2^e \in \mathcal{S}$ is derived to align the block's dynamic range with the representable limits of the element data type $\mathcal{E}$. Let $q_{\max}$ denote the maximum representable magnitude of $\mathcal{E}$ (e.g., $q_{\max} = 6.0$ for the E2M1 configuration of MXFP4). The exponent $e$ is determined by rounding the required scaling factor to the upward-adjacent power-of-two, constrained by the bit-width of the E8M0 format:
\begin{equation}
e = \text{clip}\left( \left\lceil \log_2 \frac{\max_i |x_i|}{q_{\max}} \right\rceil, -127, 127 \right),
\end{equation}
where $\text{clip}(v, a, b) = \max(a, \min(v, b))$. Subsequently, the individual elements $x_i$ are normalized and mapped to the codewords $q_i \in \mathcal{E}$ via:
\begin{equation}
    q_i = \text{round}_{\mathcal{E}}\left( \text{clip} \left( \frac{x_i}{s}, -q_{\max}, q_{\max} \right) \right).
\end{equation}
Here, $\text{round}_{\mathcal{E}}(\cdot)$ denotes the rounding operator onto the set of representable values in $\mathcal{E}$, typically utilizing round-to-nearest (RTN) logic. The dequantized reconstruction is subsequently recovered as $\hat{x}_i = 2^e\cdot q_i$.

\section{NVFP4 Transformation Logic}\label{sec:to_nvfp}
The element data type for NVFP4 is defined as E2M1, as summarized in Table~\ref{tab:nvfp4-configs}.

Mapping a tensor $\mathcal{X}  \in \mathbb{R}^{d_1 \times d_2 \times \dots \times d_n}$ in high-precision to the NVFP4 domain involves a hierarchical three-stage transformation: per-tensor scale estimation, per-block scale estimation, and element-wise quantization. First, a per-tensor scale $s_2$ is derived to normalize the overall dynamic range:
\begin{equation}
s_2 = \frac{\max(|\mathcal{X}|)}{v_{\max}},
\end{equation}
where $\ v_{\max} = \text{E4M3}_{\max} \times \text{E2M1}_{\max} = 448 \times 6$.
The tensor is pre-scaled as $\tilde{\mathcal{X}} = \mathcal{X} / s_2$. Then, for each micro-block $\tilde{\mathbf{x}} \in [-v_{\max}, v_{\max}]^k$ within $\tilde{\mathcal{X}}$, a per-block scale $s_1$ is calculated and projected onto the E4M3 codeword set:
\begin{equation}
    s_1 = \text{round}_{\text{E4M3}}\left(\frac{\max_{i}|\tilde{x}_i|}{6}\right).
\end{equation}
Finally, the normalized elements are mapped to the E2M1 codewords via:
\begin{equation}
    q_i = \text{round}_{\text{E2M1}}\left(\frac{\tilde{x}_i}{{s}_1}\right).
\end{equation}
The dequantized reconstruction is subsequently recovered as $\hat{x}_i = s_1\cdot {s}_2 \cdot q_i$. By introducing a high-precision FP32 per-tensor scaling factor $s_2$ calibrated to $v_{\max}$, the NVFP4 format theoretically eliminates clipping errors during the quantization process. Specifically, $s_2$ ensures the pre-scaled elements $\tilde{x}_i$ are within the joint representable range of the E4M3 micro-scales and E2M1 codewords, while $s_1$ subsequently maps the local distribution to the dynamic range of the E2M1 elements. Consequently, the quantization noise in NVFP4 is dominated by rounding error, which arises from the projection of values onto the discrete grids of $\mathcal{S}_1$ and $\mathcal{E}$.

\begin{table}[!t]
  \centering
  \renewcommand{\arraystretch}{1.0}
  \setlength{\tabcolsep}{8pt}
  \resizebox{\linewidth}{!}{
  \begin{tabular}{@{\hspace{1em}}l c c c@{\hspace{1em}}}
    \toprule
    \textbf{Name} & \textbf{Block Size ($k$)} & \textbf{Scale ($\mathcal{S}$)} & \textbf{Element ($\mathcal{E}$)}  \\
    \midrule
    \textbf{MXFP8}   & 32 & E8M0 & E5M2 / E4M3\\
    \textbf{MXFP6}   & 32 & E8M0 & E3M2 / E2M3 \\
    \textbf{MXFP4}   & 32 & E8M0 & E2M1 \\
    \textbf{MXINT8}  & 32 & E8M0 & INT8 \\
    \bottomrule
  \end{tabular}
  }
  \caption{Microscaling (MX) Configurations.}
  \label{tab:mx-configs}
\end{table}

\begin{table}[!t]
  \centering
  \renewcommand{\arraystretch}{1.0}
  \setlength{\tabcolsep}{24pt}
  \resizebox{0.7\linewidth}{!}{
  \begin{tabular}{@{\hspace{1em}}l l@{\hspace{1em}}}
    \toprule
   \textbf{Configurations} & \textbf{Value}\\
    \midrule
    \textbf{Block Size} ($k$)                & 16 \\
    \textbf{Per-block Scale} ($\mathcal{S}_1$) & E4M3 \\
    \textbf{Per-tensor Scale} ($\mathcal{S}_2$) & FP32 \\
    \textbf{Element} ($\mathcal{E}$)         & E2M1 \\
    \bottomrule
  \end{tabular}
  }
  \caption{NVFP4 Configurations.}\label{tab:nvfp4-configs}
\end{table}

\section{HiF8 Transformation Logic}\label{sec:to_hif8}
To transform a high-precision input $x \in \mathbb{R}$ into HiF8, the process executes a three-part pipeline. First, the binary exponent $e$ of the input magnitude is determined by the floor of the base-2 logarithm:
\begin{equation}
e = \lfloor \log_2 (|x| + \epsilon) \rfloor,
\end{equation}
where $\epsilon$ is a small constant (e.g., $2^{-14}$ for FP16 or $2^{-45}$ for FP32) introduced for numerical stability and to define the underflow boundary. This $e$ serves as the scaling factor that determines the dynamic range interval $[2^e, 2^{e+1})$ in which $|x|$ resides.

Unlike static floating-point formats, HiF8 utilizes a magnitude-dependent precision allocation. The quantity of mantissa bits $n_m$ is governed by the magnitude of the exponent $|e|$ in accordance with the bit layout design presented in Figure~\ref{fig:hif8}. This relationship is formally characterized by the following definition:
\begin{equation}
n_m =
\begin{cases}
3, & \text{if } |e| \leq 3; \\
2, & \text{if } 3 < |e| \leq 7; \\
1, & \text{if } 7 < |e| \leq 15; \\
0, & \text{if } |e| > 15.
\end{cases}
\end{equation}
The mantissa value $m$ represents the fractional component of the value, such that $|x| = (1 + m) \cdot 2^e$ for normal numbers. 

To discretize $m$ using $n_m$ bits, we scale $|x|$ by the quantization step size $\Delta = 2^{e-n_m}$:
\begin{equation}
\tilde{x} = \frac{|x|}{2^{e-n_m}}.
\end{equation}
We then apply the RTN logic to find the nearest discrete state $\hat{x}$:
\begin{equation}
\hat{x} = \lfloor \tilde{x} + 0.5 \rfloor.
\end{equation}
The quantized mantissa value is reconstructed as $1 + m_q = \hat{x} \cdot 2^{-n_m}$, where $m_q$ represents the fractional mantissa stored in the bit-field.

Following Equation~\ref{equ:hif8_normal}, the reconstructed value $x_{\text{hif8}}$ is obtained by projecting the encoding back to the floating-point domain and applying the sign bit $(-1)^s = \text{sgn}(x)$:
\begin{equation}
x_{\text{hif8}} = (-1)^s \cdot \hat{x} \cdot 2^{e-n_m}.
\end{equation}
The complete pseudocode for the HiF8 adaptive quantization and dequantization logic is summarized in Algorithm~\ref{alg:hif8}.

\begin{algorithm}[!t]
\SetAlgoLined
\KwIn{Input $x$, small constant $\epsilon$}
\KwOut{Quantized-dequantized $x_{\text{hif8}}$}
\BlankLine
$s \gets \text{sign}(x)$\;
$e \gets \lfloor \log_2(x_{|x|} + \epsilon) \rfloor$\;
\BlankLine
\tcp{Adaptive Mantissa Selection}
\uIf{$|e| \leq 3$}{
  $n_m \gets 3$\;
} \uElseIf{$|e| \leq 7$}{
  $n_m \gets 2$\;
} \uElseIf{$|e| \leq 15$}{
  $n_m \gets 1$\;
} \uElse{
  $n_m \gets 0$\;
}
\BlankLine
\tcp{Quantization-Dequantization}
$\hat{x} \gets \lfloor (|x| / 2^{e - n_m}) + 0.5 \rfloor$\;
$x_{\text{hif8}} \gets s \cdot (\hat{x} \cdot 2^{e - n_m})$\;
\Return $x_{\text{hif8}}$\;
\caption{HiF8 Adaptive Quantization}\label{alg:hif8}
\end{algorithm}

\section{Scaling HiF8 Quantization}\label{sec:scale_hif8}

The standard HiF8 format reserves significant representational capacity for extreme values that are rarely present in typical distributions (Figure~\ref{fig:cdf}). To address this, a scaling mechanism is proposed to better align the tensor's dynamic range with the high-density regions of the HiF8 encoding.

For a vector $\mathbf{x} \in \mathbb{R}^n$, a scaling factor $s$ is calculated as:
\begin{equation}
    s = \frac{K}{\max_i\left|\mathbf{x}_i\right|+\epsilon},
\end{equation}
where $K$ is a predefined hyperparameter representing the target maximum magnitude, and $\epsilon > 0$ is a small constant to ensure numerical stability. The vector $\mathbf{x}$ is then scaled to produce the normalized representation $\tilde{\mathbf{x}}$:
\begin{equation}
    \tilde{\mathbf{x}} = s \cdot \mathbf{x}.
\end{equation}
The normalized components of $\tilde{\mathbf{x}}$ are subsequently mapped to the nearest HiF8 representational levels. During inference, it is recovered by dividing the scaling factor back. This adjustment seeks to reduce the bit waste identified in Section~\ref{sec:hif8} and recover the precision necessary for high-fidelity compression.

\section{HiF4 Transformation Logic}\label{sec:to_hif4}
Given a tensor $\mathcal{X} \in \mathbb{R}^{d_1 \times d_2 \times \dots \times d_n}$, HiF4 quantization can be performed along any chosen dimension. Assume the $k$-th dimension is selected such that $d_k \equiv 0 \pmod{64}$, and let $b = d_k / 64$. Without loss of generality, we consider the vector $\mathbf{x} \in \mathbb{R}^{d_k}$ by fixing all indices of $\mathcal{X}$ except for the $k$-th dimension, denoted as $\mathbf{x}=\mathcal{X}_{i_1,\dots,:,\dots,i_n}$. Subsequently, HiF4 quantization is applied to each vector in a block-based hierarchical scaling scheme. 

First, the vector $\mathbf{x}$ is partitioned into blocks of size 64 by reshaping it into a tensor $\mathbf{X} \in \mathbb{R}^{b \times 8 \times 2 \times 4}$. For each block, we compute the maximum absolute value along each of the three sub-dimensions sequentially:
\begin{subequations}
\label{eq:hierarchical_scaling}
    \begin{align}
        \mathbf{A}_3 &= \max_{k} |\mathbf{X}_{h, i, j, k}| \in \mathbb{R}_+^{b \times 8 \times 2}, \\
        \mathbf{A}_2 &= \max_{j} (\mathbf{A}_3)_{h, i, j} \in \mathbb{R}_+^{b \times 8}, \\
        \mathbf{A}_1 &= \max_{i} (\mathbf{A}_2)_{h, i} \in \mathbb{R}_+^{b},
    \end{align}
\end{subequations}
where $|\cdot|$ denotes the absolute value operator. Subsequently, we derive a set of hierarchical scaling factors based on $\mathbf{A}_1$, $\mathbf{A}_2$, and $\mathbf{A}_3$, progressing from coarse to fine granularity.

For every 64 elements, we calculate a block-wise scaling factor $\mathbf{S}_1$ based on the normalized $\mathbf{A}_1$, which is then represented in the unsigned E6M2 format:
\begin{subequations}
    \begin{align}
        \tilde{\mathbf{A}}_1 &= \text{clip}\left(\frac{\mathbf{A}_1}{7}, q_{\min}, q_{\max}\right),\\
        \mathbf{E}_1 &= \left\lfloor\log_2{\tilde{\mathbf{A}}_1}\right\rfloor, \\
        \mathbf{M}_1 &= \text{round}\left(\frac{\tilde{\mathbf{A}}_1}{2^{\mathbf{E}_1-2}}\right), \\
        \mathbf{S}_1 &= \mathbf{M}_1 \times 2^{\mathbf{E}_1-2},
    \end{align}
\end{subequations}
where $q_{\min}=2^{-48}$ and $q_{\max}=1.5\times 2^{15}$ are the extreme values of the E6M2 format. Within each block, a sub-block scaling factor $\mathbf{S}_2$ is calculated every 8 elements based on $\mathbf{A}_2$ and $\mathbf{S}_1$. This factor is represented as a 1-bit exponent according to the following logic:
\begin{subequations}
    \begin{align}
        \tilde{\mathbf{A}}_2 &= \text{clip}\left(\frac{\mathbf{A}_2}{\mathbf{S}_1}, 0, 4\right), \\
        \mathbf{E}_2 &= \left\lfloor\frac{\tilde{\mathbf{A}}_2}{4}\right\rfloor \in \{0, 1\}, \\
        \mathbf{S}_2 &= 2^{\mathbf{E}_2}.
    \end{align}
\end{subequations}

Within each sub-block, a scaling factor $\mathbf{S}_3$ is calculated every 4 elements based on $\mathbf{A}_3$, $\mathbf{S}_1$, and $\mathbf{S}_2$. This factor is also represented as a 1-bit exponent according to the following logic:
\begin{subequations}
    \begin{align}
        \tilde{\mathbf{A}}_3 &= \text{clip}\left(\frac{\mathbf{A}_3}{\mathbf{S}_1 \times \mathbf{S}_2}, 0, 2\right), \\
        \mathbf{E}_3 &= \left\lfloor\frac{\tilde{\mathbf{A}}_3}{2}\right\rfloor \in \{0, 1\}, \\
        \mathbf{S}_3 &= 2^{\mathbf{E}_3}.
    \end{align}
\end{subequations}

Finally, the magnitude of each element is normalized by the hierarchical scaling factors $\mathbf{S}_1$, $\mathbf{S}_2$, and $\mathbf{S}_3$. The resulting values are then quantized into the E1M2 format as follows:
\begin{subequations}
    \begin{align}
        \tilde{\mathbf{X}} &= \text{clip}\left(\frac{|\mathbf{X}|}{\mathbf{S}_1 \times \mathbf{S}_2 \times \mathbf{S}_3}, 0, 1.75\right),\\
        \hat{\mathbf{X}} &= \left\lfloor \tilde{\mathbf{X}} \times 2^{2} + 0.5 \right\rfloor \in \{0, 1, \dots, 7\}.
    \end{align}
\end{subequations}

In summary, each element of $\mathbf{X}$ is represented by the tuple $(\text{sign}, \mathbf{E}_1, \mathbf{M}_1, \mathbf{E}_2, \mathbf{E}_3, \hat{\mathbf{X}})$. The corresponding dequantized value $\mathbf{X}_{\text{hif4}}$ is reconstructed as follows:
\begin{equation}
    \mathbf{X}_\text{hif4} = \text{sign} \times \mathbf{M}_1 \times 2^{\mathbf{E}_1+\mathbf{E}_2+\mathbf{E}_3-4} \times \hat{\mathbf{X}}.
\end{equation}
This formulation allows the hardware to reconstruct the value using simple integer addition of the exponent components ($\mathbf{E}_1, \mathbf{E}_2, \mathbf{E}_3$) followed by a bit-shift, avoiding the need for expensive floating-point multiplications for the scaling factors.

\section{Quantization SQNR Comparisons}
\begin{figure*}[!ht]
    \centering
    \pagebreak[0] 
    \vspace{1em}

    \captionsetup[subfigure]{font=tiny, skip=1pt} 
    \setlength{\tabcolsep}{1pt} 
    
    \foreach \i in {0,1,...,35}{%
        \begin{subfigure}{0.23\textwidth}
            \centering
            \includegraphics[width=\textwidth]{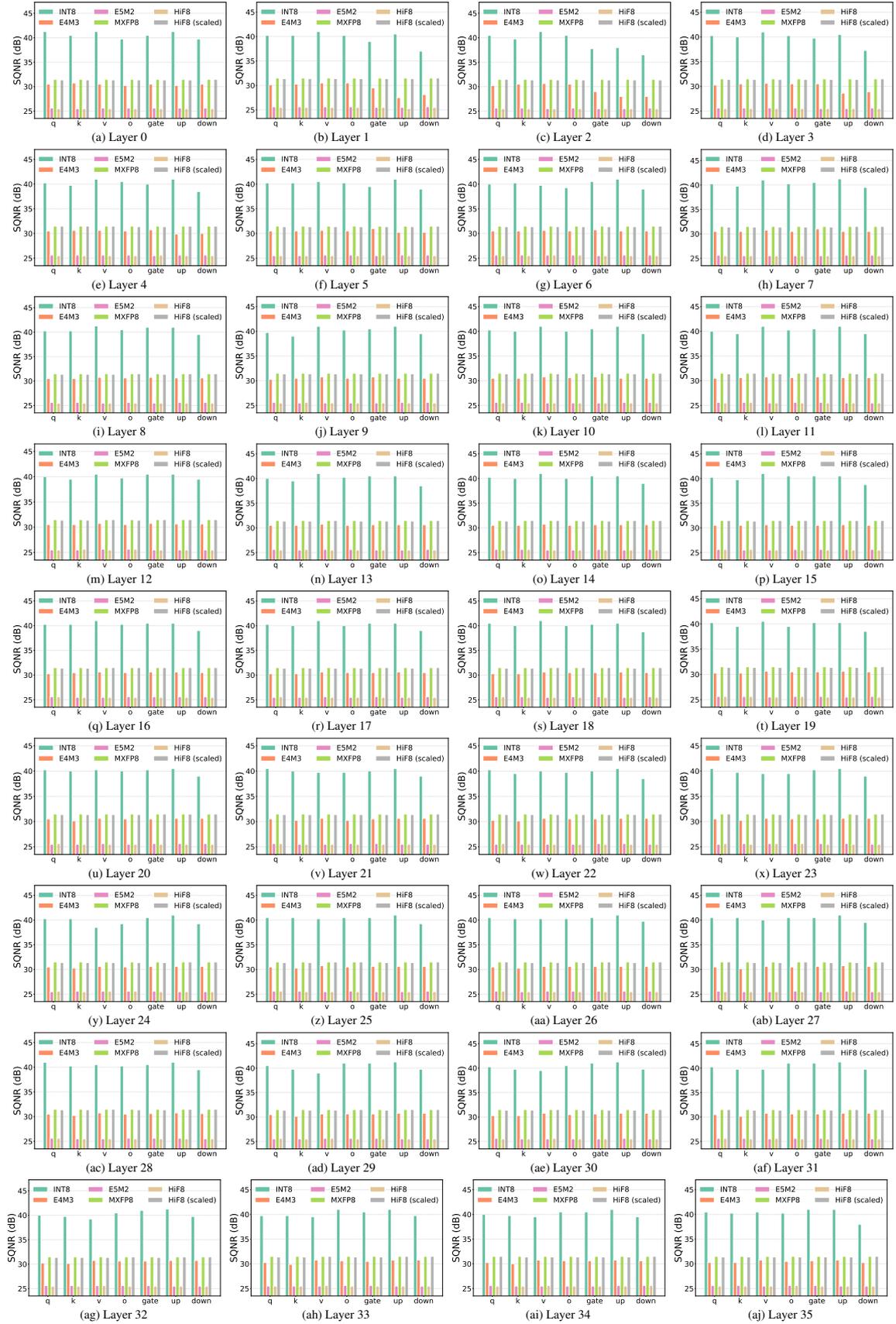}
            \caption{Layer \i} 
        \end{subfigure}
    }
    \vspace{1em}
    \caption{Weight Quantization SQNR Comparison for Qwen3-8B across 8-bit Formats.}\label{fig:qsnr_8bit_w_all}
\end{figure*}

\begin{figure*}[!ht]
    \centering
    \pagebreak[0] 
    \vspace{1em}

    \captionsetup[subfigure]{font=tiny, skip=1pt} 
    \setlength{\tabcolsep}{1pt} 
    
    \foreach \i in {0,1,...,35}{%
        \begin{subfigure}{0.23\textwidth}
            \centering
            \includegraphics[width=\textwidth]{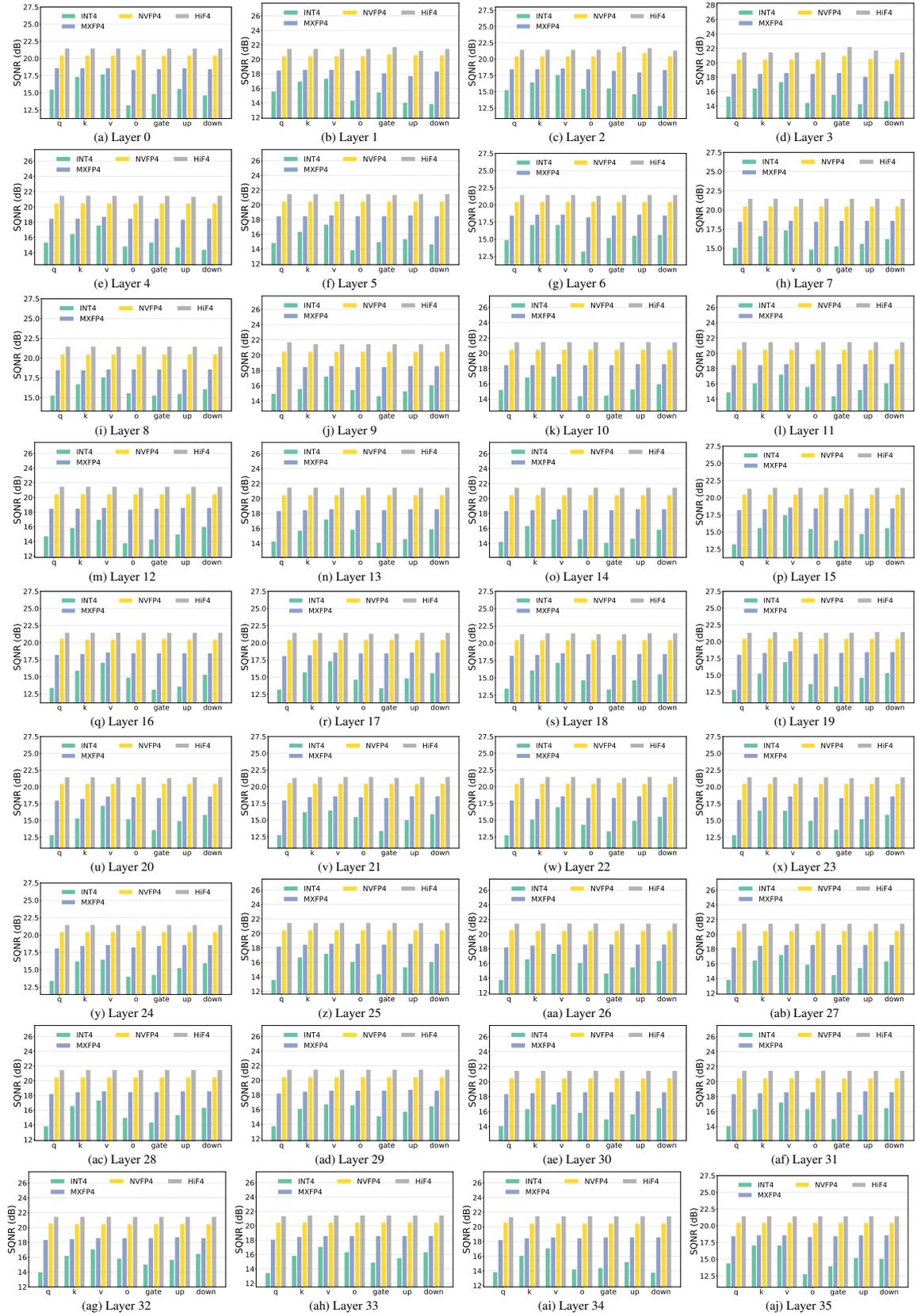}
            \caption{Layer \i} 
        \end{subfigure}
    }
    \vspace{1em}
    \caption{Weight Quantization SQNR Comparison for Qwen3-8B across 4-bit Formats.}\label{fig:qsnr_4bit_w_all}
\end{figure*}

\begin{figure*}[!ht]
    \centering
    \pagebreak[0] 
    \vspace{1em}

    \captionsetup[subfigure]{font=tiny, skip=1pt} 
    \setlength{\tabcolsep}{1pt} 
    
    \foreach \i in {0,1,...,35}{%
        \begin{subfigure}{0.23\textwidth}
            \centering
            \includegraphics[width=\textwidth]{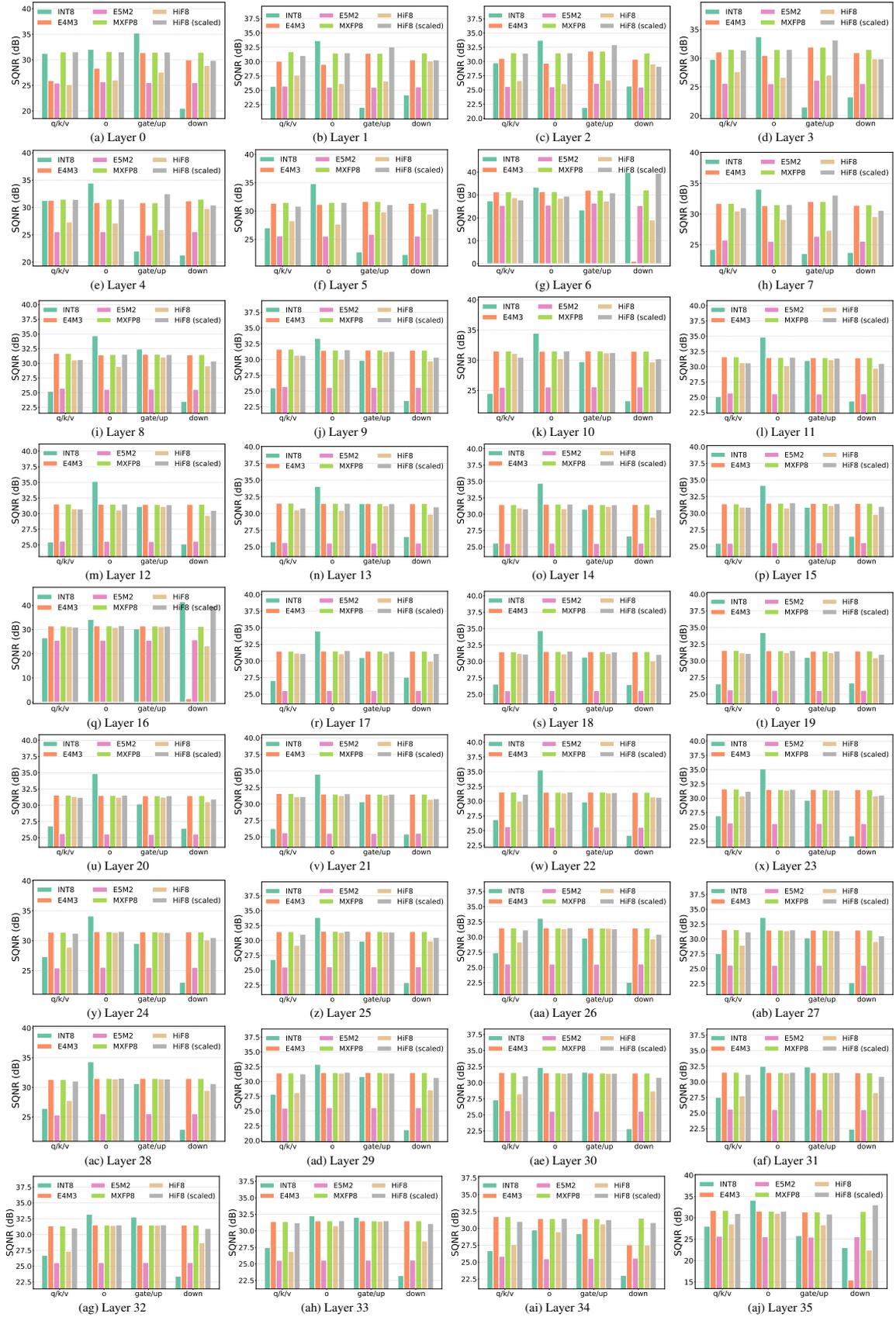}
            \caption{Layer \i} 
        \end{subfigure}
    }
    \vspace{1em}
    \caption{Activation Quantization SQNR Comparison for Qwen3-8B across 8-bit Formats.}\label{fig:qsnr_8bit_a_all}
\end{figure*}

\begin{figure*}[!ht]
    \centering
    \pagebreak[0] 
    \vspace{1em}

    \captionsetup[subfigure]{font=tiny, skip=1pt} 
    \setlength{\tabcolsep}{1pt} 
    
    \foreach \i in {0,1,...,35}{%
        \begin{subfigure}{0.23\textwidth}
            \centering
            \includegraphics[width=\textwidth]{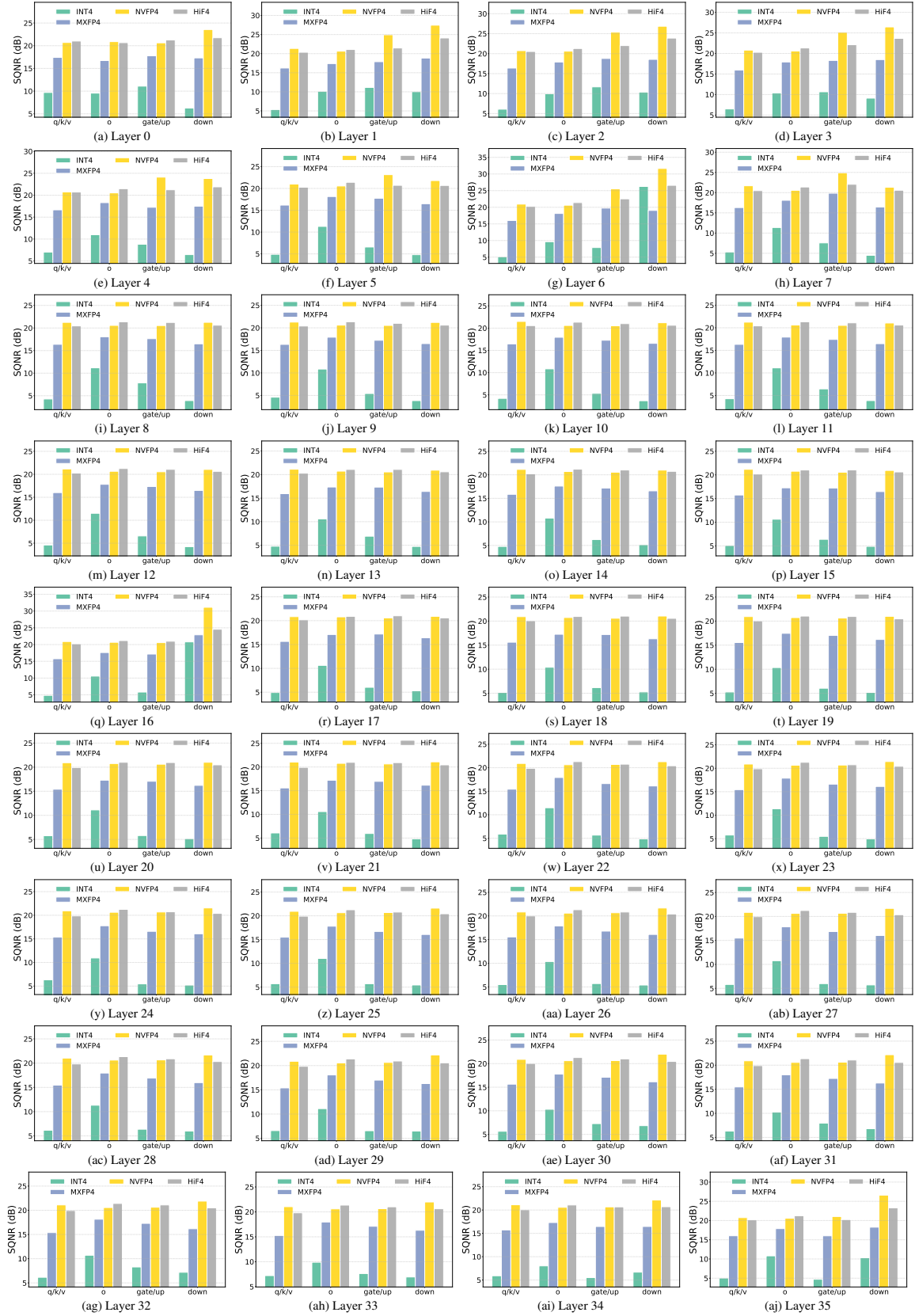}
            \caption{Layer \i} 
        \end{subfigure}
    }
    \vspace{1em}
    \caption{Activation Quantization SQNR Comparison for Qwen3-8B across 4-bit Formats.}\label{fig:qsnr_4bit_a_all}
\end{figure*}

\subsection{Weight Quantization}\label{appendix:sqnr_weight}
We evaluate the impact of reduced precision on model fidelity by analyzing the layer-wise SQNR for Qwen3-8B across all 36 decoder layers. Figures~\ref{fig:qsnr_8bit_w_all} and ~\ref{fig:qsnr_4bit_w_all} contrast the representational efficiency of investigated formats in 8-bit and 4-bit regimes.

As shown in Figure~\ref{fig:qsnr_8bit_w_all}, in the 8-bit regime, all formats maintain high fidelity. INT8 consistently achieves the highest SQNR, outperforming floating-point formats by approximately 5--10 dB. This indicates that for 8-bit weights, the uniform density of an integer format is more effective than an expansive dynamic range. The bit-waste incurred by exponent bits in MXFP8 or HiF8 results in a coarser representation for the bulk of the weight distribution compared to the fine-grained, uniform steps of INT8. By employing the per-channel scaling mechanism detailed in Appendix~\ref{sec:scale_hif8} with $K=16$, HiF8 (scaled) achieves performance comparable to that of MXFP8.

As illustrated in Figure~\ref{fig:qsnr_4bit_w_all}, the transition to 4-bit precision marks a shift where uniform spacing becomes a bottleneck, causing INT4 SQNR to collapse, often falling below 15 dB. In this constrained regime, the hierarchical structure of HiF4 and NVFP4 proves essential. HiF4 consistently yields the highest SQNR across all components ($W_q, W_k, W_v, W_o$, and MLP layers). By utilizing multi-level scaling to isolate outliers while preserving local precision, HiF4 demonstrates superior structural resilience, validating its efficacy for high-fidelity inference at ultra-low bit-widths.

\begin{figure*}[!t]
    \centering
    \pagebreak[0] 
    \vspace{-2em}
    \begin{subfigure}{0.40\textwidth}
        \centering
            \includegraphics[width=\textwidth, trim=0 0.1cm 0 4cm, clip]{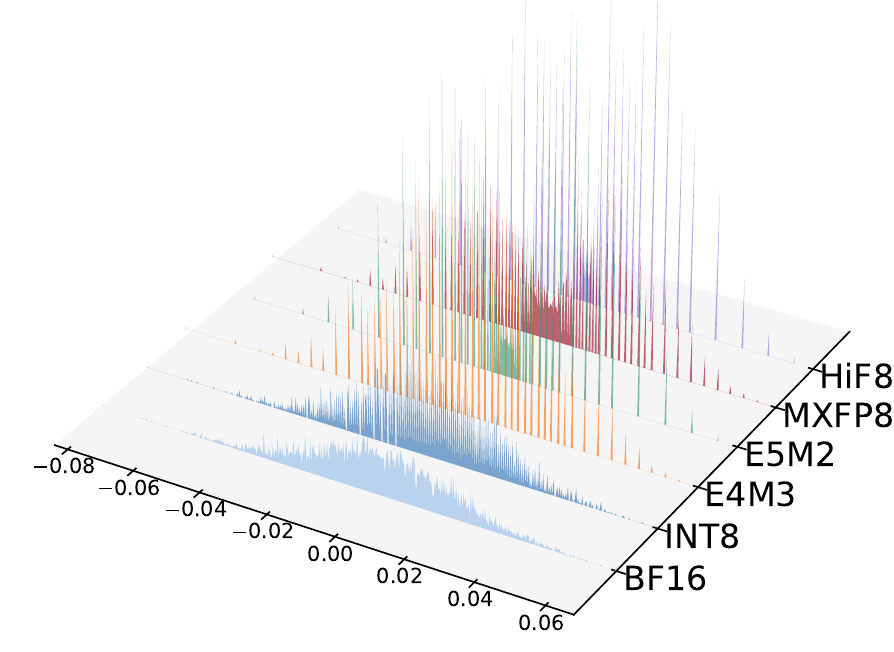}
            \caption{Weight of Layer 3 $\texttt{down\_proj}$} \label{fig:dist_8bit_w}
    \end{subfigure}
    \begin{subfigure}{0.40\textwidth}
        \centering
            \includegraphics[width=\textwidth, trim=0 0.1cm 0 4cm, clip]{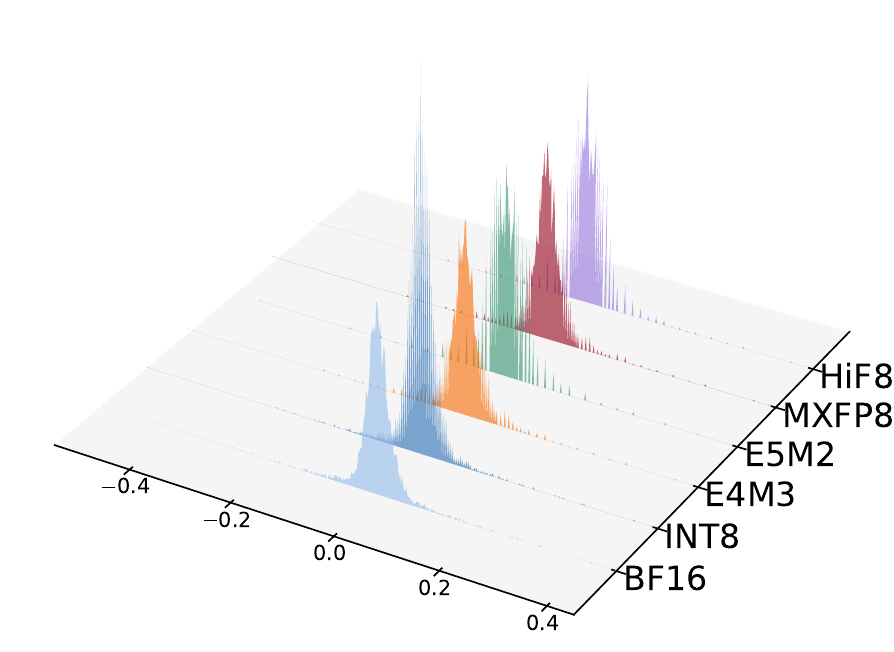}
            \caption{Activation of Layer 3 $\texttt{q/k/v\_proj}$} \label{fig:dist_8bit_a}
    \end{subfigure} \\
    \begin{subfigure}{0.40\textwidth}
        \centering
            \includegraphics[width=\textwidth, trim=0 0.1cm 0 4cm, clip]{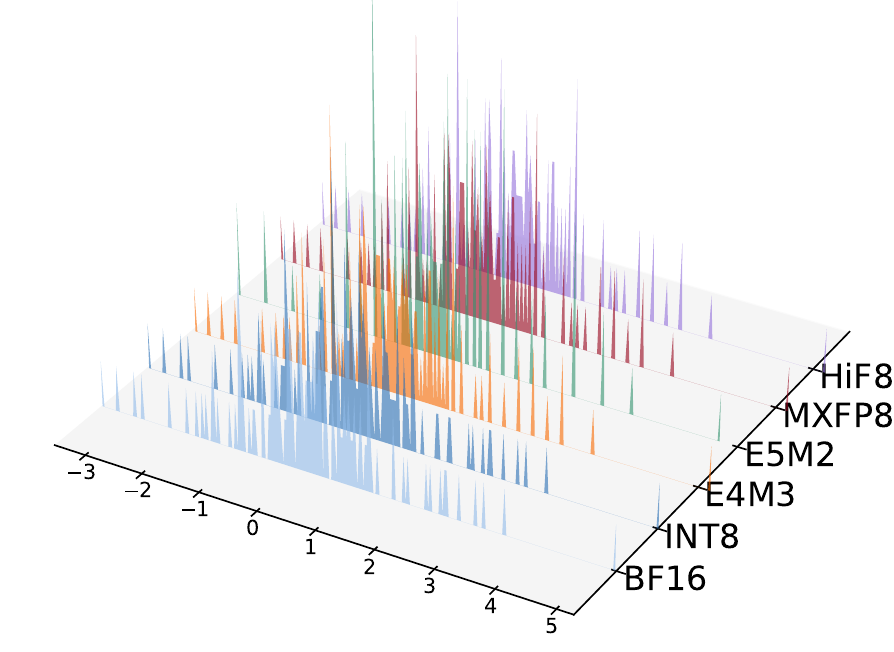}
            \caption{Key State of Layer 3} \label{fig:dist_8bit_k}
    \end{subfigure}
    \begin{subfigure}{0.40\textwidth}
        \centering
            \includegraphics[width=\textwidth, trim=0 0.1cm 0 4cm, clip]{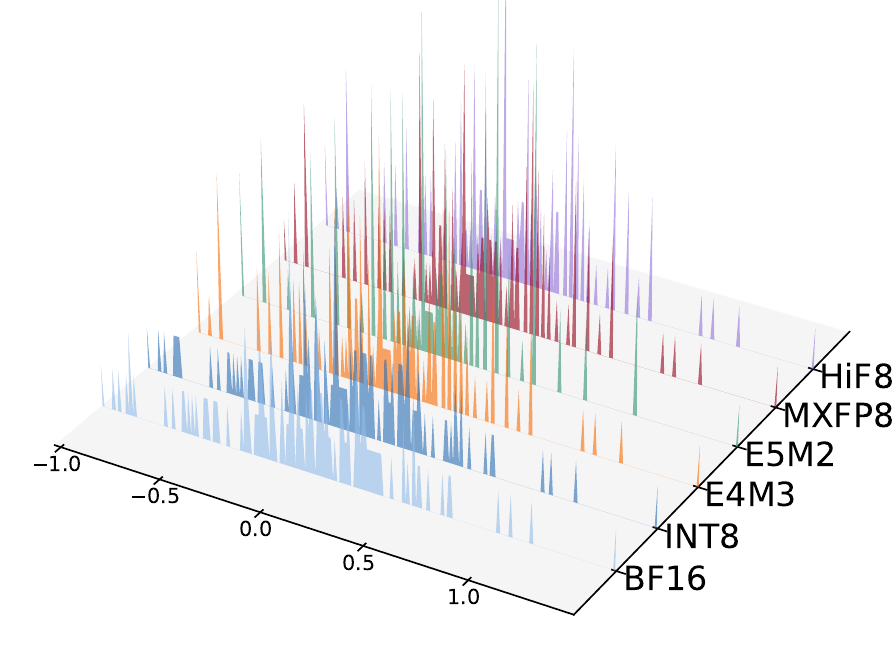}
            \caption{Value State of Layer 3} \label{fig:dist_8bit_v}
    \end{subfigure}
    \caption{Distributional Impact of 8-bit Quantization Formats on Qwen3-8B Layer 3 Components ($z$-axis: density).}
\end{figure*}

\subsection{Activation Quantization}\label{appendix:sqnr_activation}
To assess the impact of reduced precision on model activations, we evaluate the layer-wise SQNR for Qwen3-8B with Wikitext-2 dataset. Unlike weights, activation distributions are dynamic and often characterized by prominent outliers, making them significantly more sensitive to quantization bit-widths. Figures~\ref{fig:qsnr_8bit_a_all} and~\ref{fig:qsnr_4bit_a_all} illustrate the activation fidelity for the 8-bit and 4-bit regimes, respectively.

As shown in Figure~\ref{fig:qsnr_8bit_a_all}, in the 8-bit regime, the performance hierarchy differs markedly from the weight analysis. While INT8 remains competitive in certain layers, floating-point formats, specifically HiF8 and MXFP8, demonstrate superior stability across the 36 decoder layers. Across most components ($W_q, W_k, W_v, W_o$, and MLP layers), HiF8 and MXFP8 maintain SQNR levels between 30 dB and 35 dB. The expanded dynamic range afforded by the exponent bits in these formats allows them to better accommodate the outlier spikes common in LLM activations. With $K=4$, HiF8 (scaled) achieves performance comparable to that of MXFP8. In contrast, INT8 shows higher variability, particularly in deeper layers where activation magnitudes fluctuate more aggressively, leading to sporadic drops in fidelity compared to the more consistent floating-point baselines.

As illustrated in Figure~\ref{fig:qsnr_4bit_a_all}, the transition to 4-bit precision reveals a near-total collapse for uniform integer quantization. INT4 SQNR values plummet below 10 dB in nearly all layers, and in several instances (such as the activations of $W_q$, $W_k$, and $W_v$ in Layer 1-3), they drop to 5 dB. This confirms that the constraint bit budget and uniform spacing of INT4 are fundamentally incapable of representing the high-variance distributions of activations. In this low precision setting, hierarchical formats become mandatory for functional integrity. HiF4 and NVFP4 consistently achieve the highest SQNR.

\section{Distributional Analysis}

\begin{figure*}[!t]
    \centering
    \pagebreak[0] 
    \vspace{-2em}
    \begin{subfigure}{0.40\textwidth}
        \centering
            \includegraphics[width=\textwidth, trim=0 0.1cm 0 4cm, clip]{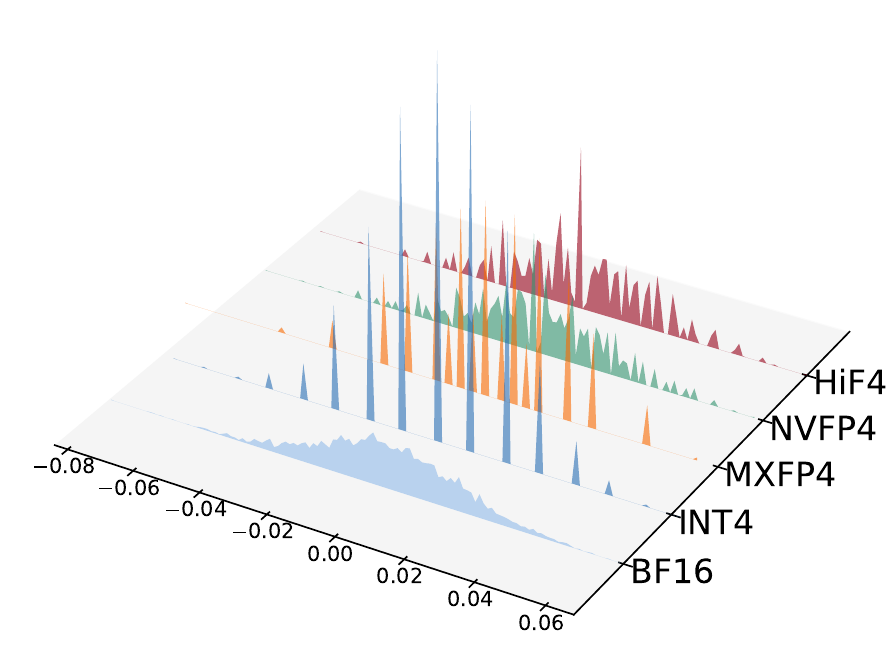}
            \caption{Weight of Layer 3 $\texttt{down\_proj}$} \label{fig:dist_4bit_w}
    \end{subfigure}
    \begin{subfigure}{0.40\textwidth}
        \centering
            \includegraphics[width=\textwidth, trim=0 0.1cm 0 4cm, clip]{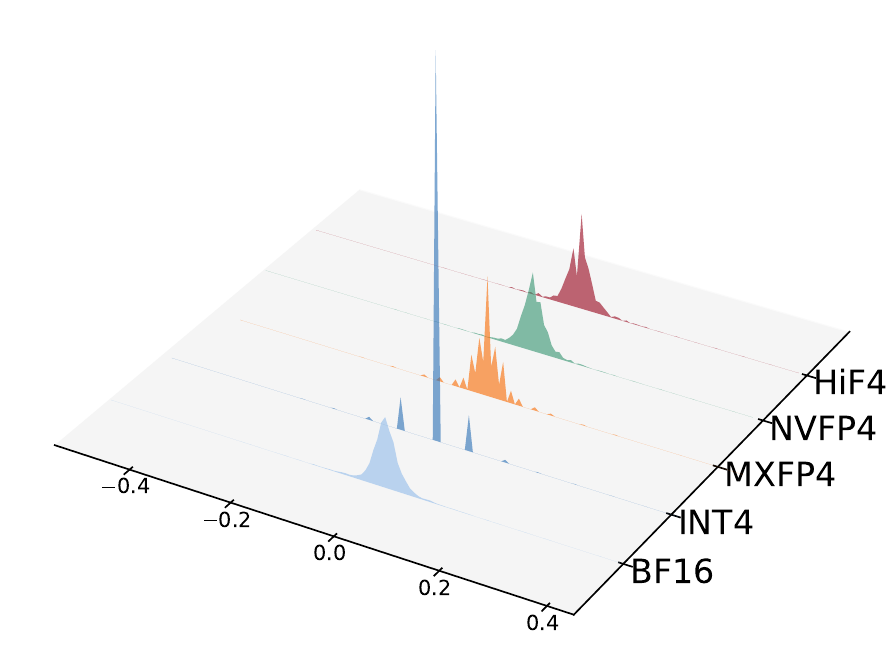}
            \caption{Activation of Layer 3 $\texttt{q/k/v\_proj}$} \label{fig:dist_4bit_a}
    \end{subfigure}\\
    \begin{subfigure}{0.40\textwidth}
        \centering
            \includegraphics[width=\textwidth, trim=0 0.1cm 0 4cm, clip]{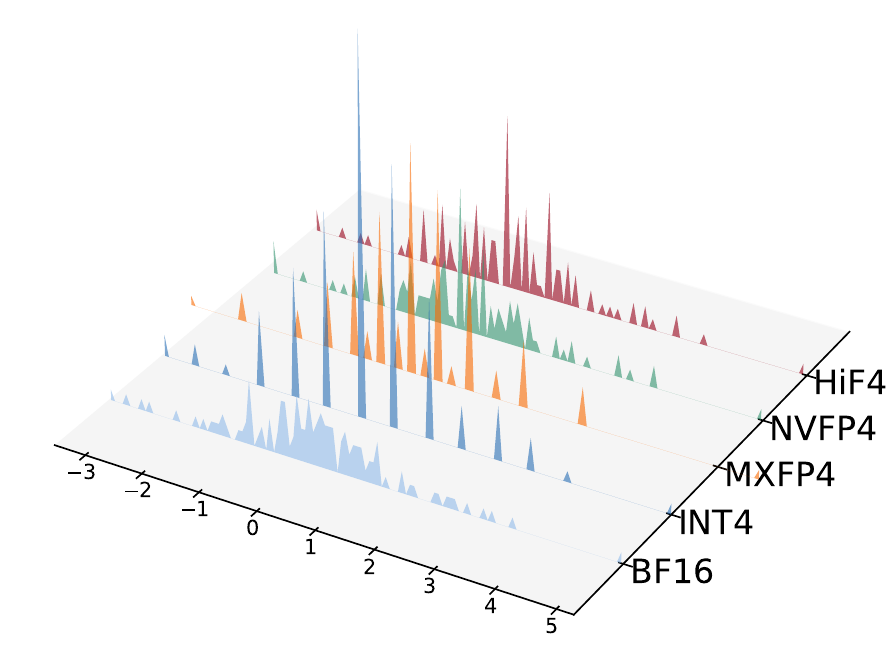}
            \caption{Key State of Layer 3} \label{fig:dist_4bit_k}
    \end{subfigure}
    \begin{subfigure}{0.40\textwidth}
        \centering
            \includegraphics[width=\textwidth, trim=0 0.1cm 0 4cm, clip]{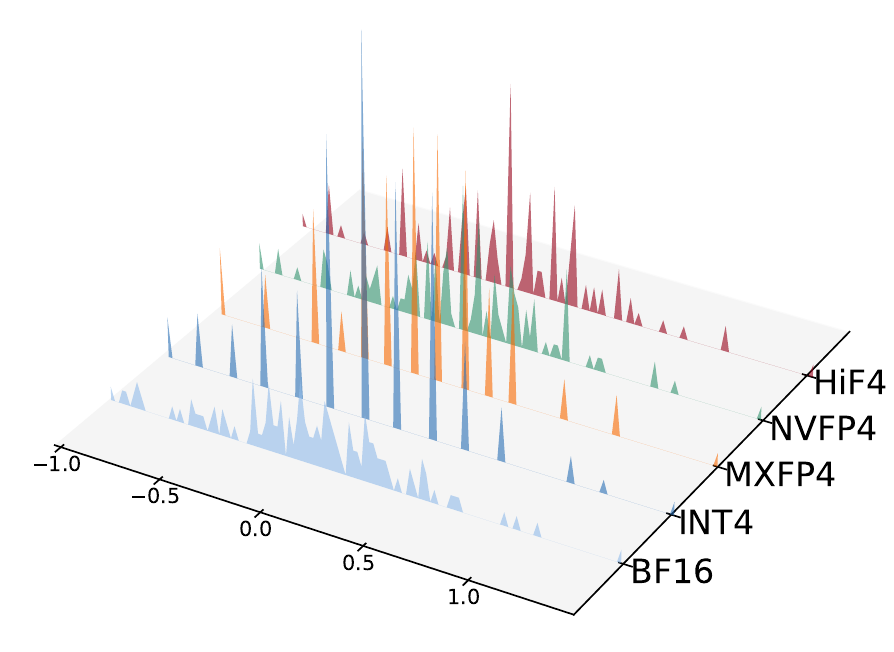}
            \caption{Value State of Layer 3} \label{fig:dist_4bit_v}
    \end{subfigure}
    \caption{Distributional Impact of 4-bit Quantization Formats on Qwen3-8B Layer 3 Components ($z$-axis: density).}
\end{figure*}

\begin{figure*}
    \centering
    \includegraphics[width=\linewidth]{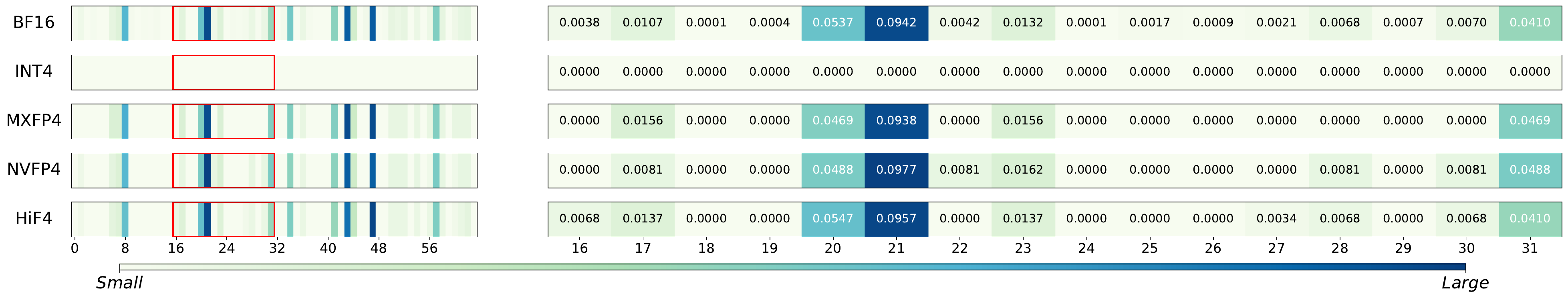}
    \caption{Visualization of Quantization Resolution within Blocks across Various 4-bit Formats. Left: Global overview of the 64-element activation vector, with the red box indicating the zoomed sub-region. Right: Element-wise magnification of indices 16--31.}
    \label{fig:block_density}
\end{figure*}

\subsection{8-bit Formats}\label{sec:dist_analysis_8bit}
Figure~\ref{fig:dist_8bit_w} illustrates the first channel of the Qwen3-8B Layer 3 $\texttt{down\_proj}$ weight distribution, which is characterized by a symmetric, bell-shaped density concentrated in a narrow range. Within this bounded space, INT8 provides the most faithful reconstruction because its uniform spacing maximizes representational density where the signal is most prevalent. In contrast, MXFP8 and HiF8 allocate a portion of their bit-budget to exponent ranges that remain largely unpopulated by weight values. This exponent-waste results in a coarser quantization grid near the center of the distribution, explaining why floating-point formats exhibit higher quantization noise for static weights.

The activation input to Layer 3 $\texttt{q/k/v\_proj}$ (Figure~\ref{fig:dist_8bit_a}) presents a sharp contrast, exhibiting significant asymmetry and heavy-tailed outliers. While INT8 utilizes a zero-point to address the global shift, its uniform nature forces a massive increase in step size to encompass extreme outliers, leading to a resolution collapse near zero. Conversely, the logarithmic spacing of MXFP8 and HiF8 allows the grid to maintain high precision in the dense region near the origin while simultaneously providing the dynamic range necessary to capture high-magnitude activations. This adaptability confirms that for dynamic activations, the structural flexibility of floating-point formats outweighs the raw density of integer formats.

Figures~\ref{fig:dist_8bit_k} and~\ref{fig:dist_8bit_v} visualize the contrasting behaviors of Key and Value tensors. The Key states display a complex, multi-modal distribution where the hierarchical flexibility of HiF8 and MXFP8 is essential for capturing shifting peaks. The Value states appear more stable and bounded, similar to weight distributions but with slightly more variance. This stability explains why the performance gap between uniform and hierarchical formats is narrower for Keys than for Values; when a distribution is bounded, the expansive range of floating formats is less critical, allowing denser uniform formats to remain highly competitive.

\begin{table*}[!th]
\centering
\small
\renewcommand{\arraystretch}{1.1}
\setlength{\tabcolsep}{6pt}
\resizebox{0.85\linewidth}{!}{
\begin{tabular}{c|c|c|c|cc|cccc|c|r}
\toprule
\multirow{2}{*}{\textbf{Models}} & \multirow{2}{*}{\textbf{\makecell{SmoothQuant \\ WA Precision}}} & \multirow{2}{*}{\textbf{\makecell{Data \\ Format}}} & \multirow{2}{*}{\textbf{\makecell{QKV Quant. \\ Precision}}} & \multicolumn{2}{c|}{\textbf{PPL} ($\downarrow$)} & \multicolumn{4}{c|}{\textbf{Acc} ($\uparrow$)} & \multirow{2}{*}{\makecell{\textbf{EM} ($\uparrow$)\\\textbf{GSM8K}}} & \multirow{2}{*}{\textbf{\makecell{Accuracy\\Loss $\Delta$(\%)$\downarrow$}}}  \\
\cline{5-10}
& & & & \textbf{Wikitext} & \textbf{C4} & \textbf{HellaSwag} & \textbf{MMLU} & \textbf{Arc-C} & \textbf{MATH500} &  &  \\
\midrule
\multirow{20}{*}{\textbf{Qwen3-8B}} & - &  BF16 & - & 9.72 & 15.43 & 0.5711 & 0.7302 & 0.5546 & 78.4 & 0.8787 & - \\ \cline{2-12}
  
  & \multirow{9}{*}{W8A8} & \multirow{3}{*}{INT8} & - & 9.59 & 15.29 & 0.5720 & 0.7276 & 0.5478  &  79.2 & 0.0334  & 13.48 \\
  & & & Q16KV8 & 9.58 & 15.29 & 0.5724 & 0.7292 & 0.5503 & 77.4 &  0.0326  & 13.70 \\
  & & & QKV8 & 9.59 & 15.30 & 0.5712 & 0.7287 & 0.5563  & 80.2 & 0.0326  & 13.10 \\ \cline{3-12}
  
  & & \multirow{3}{*}{MXFP8} & - & 9.75 & 15.48 & 0.5695 & 0.7277 & 0.5606 & 78.6 &  0.8825  & \textcolor{red}{\textbf{-0.07}}\\
  & & & Q16KV8 & 9.77 & 15.48 & 0.5696 & 0.7264 & 0.5538 & 79.8  &   0.8832  & \textcolor{red}{\textbf{-0.08}} \\
  & & & QKV8 & 9.77 & 15.49 & 0.5707 & 0.7265 & 0.5580 & 79.0 &  0.8886  & \textcolor{red}{\textbf{-0.15}} \\ \cline{3-12}
  
  & & \multirow{3}{*}{HiF8} & \cc - & \cc 9.79 & \cc 15.47 & \cc 0.5680 & \cc 0.7227 & \cc 0.5538 & \cc 79.8 &   \cc 0.8787 &  \cc  \textcolor{blue}{\textit{0.13}} \\
  & & & \cc Q16KV8 & \cc 9.81 & \cc 15.50 &\cc  0.5684 &\cc 0.7258 &\cc 0.5512 &\cc  82.8 &\cc  0.8741  &\cc \textcolor{blue}{\textit{-0.29}} \\
  & & &\cc QKV8 &\cc 9.85 &\cc 15.54 &\cc 0.5658 &\cc 0.7236 &\cc 0.5435 &\cc 81.0  &\cc  0.8772  &\cc  \textcolor{blue}{\textit{0.39}} \\
  \cline{2-12}

& \multirow{12}{*}{W4A4} & \multirow{3}{*}{INT4} & -  & 6.1e2 & 6.0e2 & 0.2651 & 0.2344 & 0.1903  & 1.4 & 0.0015  & 1478.49 \\
  & & & Q16KV4 & 
  6.3e2 & 6.3e2
  & 0.2649 & 0.2333 & 0.2090  & 1.4  & 0.0008  & 1535.22 \\
  & & & QKV4 & 
  7.9e2 & 8.0e2
  & 0.2591 & 0.2415 & 0.1937  & 0.8 & 0.0008  & 1928.26 \\ \cline{3-12}
  
  & & \multirow{3}{*}{MXFP4} & - & 10.90 & 17.46 & 0.5352 & 0.6700 & 0.4923  & 74.8 & 0.8188  &  8.92 \\
  & & & Q16KV4 & 11.08 & 17.67 & 0.5267 & 0.6587 & 0.4821  & 74.2 & 0.7688  & 11.00 \\
  & & & QKV4 & 11.48 & 18.27 & 0.5120 & 0.6422 &  0.4761 & 68.0 &  0.7392  & 14.60 \\ \cline{3-12}
  
  & & \multirow{3}{*}{NVFP4} & - & 10.12 & 16.21 & 0.5491 & 0.7079 & 0.5324  & 80.6 &  0.8711  & \textcolor{red}{\textbf{2.59}} \\
  & & & Q16KV4 & 10.20 & 16.28 & 0.5473 & 0.7041 & 0.5205  & 81.2  & 0.8484  &  \textcolor{blue}{\textit{3.46}} \\
  & & & QKV4 & 10.28 & 16.41 & 0.5446 & 0.6988 & 0.5128  & 75.2  & 0.8522  & \textcolor{blue}{\textit{5.10}} \\ \cline{3-12}
  
  & & \multirow{3}{*}{HiF4}  & \cc - & \cc 10.27 & \cc 16.31 & \cc 0.5566 & \cc 0.7050 & \cc 0.5478 & \cc 77.2 & \cc 0.8552   & \cc \textcolor{blue}{\textit{3.25}} \\
  & & & \cc Q16KV4 &\cc 10.23 &\cc 16.31 &\cc 0.5563 &\cc 0.7051 &\cc 0.5435  & \cc 77.8 &\cc  0.8582  &\cc \textcolor{red}{\textbf{3.15}} \\
  & & & \cc QKV4 & \cc10.32 &\cc 16.47 &\cc 0.5515 &\cc 0.6939 &\cc 0.5478  & \cc 78.8 & \cc  0.8309  & \cc \textcolor{red}{\textbf{3.92}}\\  \midrule 
  
  \multirow{20}{*}{\textbf{openPangu-7B}} & - &  BF16 & - &  34.95 & 57.18 & 0.4504 & 0.5667 &  0.3225  & 89.8 & 0.5049  & - \\ \cline{2-12}
  & \multirow{9}{*}{W8A8} & \multirow{3}{*}{INT8}  & - & 35.79 & 56.93 & 0.4503 & 0.5882 & 0.3208   & 86.6 & 0.0250  & 13.90 \\ 
  & & & Q16KV8 & 35.86 & 57.11 & 0.4512 & 0.5894 & 0.3294 &  88.6 &  0.0281  & 13.13 \\
  & & & QKV8 & 35.91 & 57.04 & 0.4482 & 0.5866 & 0.3200 & 91.0 &  0.0227  & 13.49 \\ \cline{3-12}

  & & \multirow{3}{*}{MXFP8}  & - & 35.01 & 56.90 & 0.4439 & 0.5799 & 0.3319 & 88.4 & 0.5034  &  \textcolor{red}{\textbf{-0.32}} \\ 
  & & & Q16KV8 & 36.03 & 58.37 & 0.4448 & 0.5454 & 0.3140  & 89.4 & 0.5095  &  \textcolor{red}{\textbf{1.76}} \\
  & & & QKV8 & 36.31 & 58.89 & 0.4452 & 0.5467 & 0.3285  & 89.4 &  0.5140  & \textcolor{red}{\textbf{1.19}} \\  
  
  \cline{3-12}
 
  & & \multirow{3}{*}{HiF8}  & \cc- &  \cc36.67 & \cc59.38 & \cc0.4481 & \cc0.5783 & \cc0.3396  & \cc90.2 & \cc0.5178  & \cc \textcolor{blue}{\textit{-0.15}}\\ 
    & & & \cc Q16KV8 & \cc 40.23 & \cc 63.18 &\cc  0.4433 & \cc 0.5642 & \cc 0.3200 & \cc 87.0  & \cc 0.4966  & \cc \textcolor{blue}{\textit{4.74}} \\
  & & & \cc QKV8 & \cc 40.46 &\cc  63.50 & \cc 0.4433 & \cc 0.5407 & \cc 0.3140 &  \cc 85.8 &  \cc 0.5155  & \cc \textcolor{blue}{\textit{5.42}} \\  
  
  \cline{2-12}

  & \multirow{12}{*}{W4A4} & \multirow{3}{*}{INT4}  & - & 6.4e2 & 7.3e2 & 0.3124 & 0.2752 & 0.2201  & 7.2  & 0.0023  & 459.03 \\ 
  & & & Q16KV4 & 
  1.9e3 & 2.2e3
  & 0.2796 & 0.2504 & 0.2048 &  2.2 &  0.0000  & 1375.22 \\
  & & & QKV4 & 
  2.3e3  & 2.6e3
  & 0.2860 & 0.2406 & 0.2210 & 0.8 &  0.0008  & 1603.48 \\  
  
  \cline{3-12}

  & & \multirow{3}{*}{MXFP4}  & - & 45.55 & 62.67 & 0.4296 & 0.5236 & 0.3148 & 83.0 &  0.4488  & 10.46\\ 
  & & & Q16KV4 & 81.51 & 88.81 & 0.4027 & 0.3296 & 0.3063 & 46.0 &  0.1698  & 51.59 \\
  & & & QKV4 & 92.62 & 97.77 & 0.3924 & 0.3448 & 0.2927 & 35.2  &  0.1077  & 62.39 \\  
  
  \cline{3-12}

  & & \multirow{3}{*}{NVFP4}  & - & 38.68 & 58.18 & 0.4372 & 0.4855 & 0.3131 & 89.4 &  0.5140  & \textcolor{blue}{\textit{4.46}}  \\ 
  & & & Q16KV4 & 44.67 & 65.68 & 0.4332 & 0.4755 & 0.3311 & 84.8  &   0.3844  &  \textcolor{blue}{\textit{12.77}}\\
  & & & QKV4 & 45.38 & 65.38 & 0.4299 & 0.4588 & 0.3012 & 79.8 &  0.3639  & \textcolor{blue}{\textit{16.21}}  \\  
  
  \cline{3-12}

  & & \multirow{3}{*}{HiF4}  & \cc - & \cc 39.69 & \cc 60.27 & \cc 0.4434 & \cc 0.5426 & \cc 0.3055 & \cc 88.6 &   \cc 0.5262  & \cc \textcolor{red}{\textbf{3.88}} \\ 
  & & & \cc Q16KV4 &\cc  43.44 & \cc 61.43 & \cc 0.4361 & \cc 0.4563 & \cc 0.3063 & \cc 84.2  &  \cc 0.4466  & \cc \textcolor{red}{\textbf{11.03}} \\
  & & & \cc QKV4 & \cc 44.76 & \cc 61.51 &  \cc 0.4368 & \cc 0.4190 & \cc 0.2944 &  \cc 83.0 &   \cc 0.4246  & \cc \textcolor{red}{\textbf{13.84}} \\  
\bottomrule
\end{tabular}
}
\caption{Performance Evaluation of KV Cache and Attention Quantization with Various Formats based on the Weight and Activation Quantization using SmoothQuant. Q16KV8 and Q16KV4 denote KV cache quantization, while QKV8 and QKV4 simulate attention quantization. Accuracy Loss denotes the average accuracy loss compared with the BF16 baselines. \textcolor{red}{\textbf{Bold red}} and \textcolor{blue}{\textit{italic blue}} indicate the best and second best settings, respectively.}
\label{tab:results-main_qkv_attention_quant}
\end{table*}

\subsection{4-bit Formats}\label{sec:dist_analysis_4bit}
The distribution in Figure~\ref{fig:dist_4bit_w} highlights the collapse of uniform quantization at ultra-low bit-widths. Because INT4 is restricted to only 16 representable levels, it must stretch its uniform grid to cover the entire weight range, resulting in extremely coarse binning and a total loss of local feature resolution. MXFP4 offers an improvement over the integer baseline but remains limited by its reliance on power-of-two scaling and a larger 32-element block size. This increased granularity prevents the format from adapting to local variations, causing the quantization intervals to widen significantly as values deviate from zero. In contrast, the hierarchical structures of HiF4 and NVFP4 allow for a much denser concentration of representable states near the bulk of the distribution, preserving the original signal's bell-shaped contour even under stringent constraints.

\begin{table*}[!t]
\centering
\small
\renewcommand{\arraystretch}{1.1}
\setlength{\tabcolsep}{6pt}
\resizebox{0.85\linewidth}{!}{
\begin{tabular}{c|c|c|c|cccccc|r}
\toprule
\textbf{Models} & \textbf{\makecell{SmoothQuant \\ WA Precision}} & \textbf{\makecell{Data \\ Format}} & \textbf{\makecell{QKV Quant. \\ Precision}} & \textbf{\makecell{Single-Document\\QA}} & \textbf{\makecell{Multi-Document\\QA}} & \textbf{Summarization} & \textbf{\makecell{Few-shot\\Learning}} & \textbf{Synthetic} & \textbf{\makecell{Code \\Completion}} & \textbf{Average} \\
\midrule
\multirow{20}{*}{\textbf{Qwen3-8B}} & - &  BF16 & - & 42.42 & 29.50 & 23.09 & 29.99 & 67.89 & 4.40 & 33.93 \\ \cline{2-11}
  & \multirow{9}{*}{W8A8} & \multirow{3}{*}{INT8} & - & 41.92 & 30.08 & 22.67 & 30.57 & 68.43 & 4.95 & 34.10 \\
  & & & Q16KV8 & 41.88 & 30.50 & 22.64 & 30.65 & 69.47 & 4.67 & 34.31 \\
  & & & QKV8 & 41.93 & 29.63 & 22.67 & 31.77 & 68.56 & 4.70 & 34.24 \\ \cline{3-11}
  & & \multirow{3}{*}{MXFP8} & - & 42.77 & 29.95 & 22.91 & 32.20 & 68.97 & 5.29 & \textcolor{red}{\textbf{34.71}} \\
  & & & Q16KV8 & 42.64 & 29.48 & 23.07 & 33.31 & 68.78 & 5.36 & \textcolor{red}{\textbf{34.82}} \\
  & & & QKV8 & 42.56 & 29.57 & 22.64 & 31.30 & 68.72 & 4.96 & 34.30 \\ \cline{3-11}
  & & \multirow{3}{*}{HiF8} &\cc - &\cc 42.23 &\cc 29.16 &\cc 22.80 &\cc 32.67 &\cc 69.14 &\cc 3.84 &\cc 34.41 \\
  & & &\cc Q16KV8 &\cc 42.05 &\cc 28.97 &\cc 22.91 &\cc 33.83 &\cc 68.89 &\cc 3.94 &\cc 34.55 \\
  & & &\cc QKV8 &\cc 42.15 &\cc 28.49 &\cc 22.86 &\cc 33.18 &\cc 68.83 &\cc 3.55 &\cc 34.30 \\
  \cline{2-11}
  & \multirow{12}{*}{W4A4} & \multirow{3}{*}{INT4} & - & 1.17 & 1.11 & 2.13 & 0.59 & 0.91 & 7.81 & 1.83 \\
  & & & Q16KV4 & 1.14 & 1.42 & 2.45 & 0.54 & 1.24 & 7.31 & 1.93 \\
  & & & QKV4 & 1.03 & 1.12 & 2.54 & 0.51 & 1.43 & 7.75 & 1.93 \\ \cline{3-11}
  & & \multirow{3}{*}{MXFP4} & - & 40.73 & 25.07 & 22.46 & 31.88 & 63.14 & 26.96 & \textcolor{red}{\textbf{34.47}} \\
  & & & Q16KV4 & 37.95 & 23.98 & 22.17 & 24.61 & 58.65 & 25.59 & 31.52\\
  & & & QKV4 & 36.20 & 23.88 & 22.27 & 22.66 & 53.72 & 27.24 & 30.27 \\ \cline{3-11}
  & & \multirow{3}{*}{NVFP4} & - &  39.34 & 28.61 & 23.00 & 25.42 & 65.10 & 8.44 & 32.27 \\
  & & & Q16KV4 & 39.25 & 26.39 & 22.93 & 26.01 & 61.97 & 7.87 & 31.43 \\
  & & & QKV4 & 38.74 & 25.88 & 22.72 & 25.95 & 61.39 & 9.83 & \textcolor{blue}{\textit{31.28}} \\ \cline{3-11}
  & & \multirow{3}{*}{HiF4} &\cc - &\cc 39.70 &\cc 27.92 &\cc 22.63 &\cc 26.77 &\cc 67.50 &\cc 5.85 &\cc \textcolor{blue}{\textit{32.49}} \\
  & & &\cc Q16KV4 &\cc 41.08 &\cc 27.91 &\cc 22.82 &\cc 28.37 &\cc 66.14 &\cc 7.34 &\cc \textcolor{red}{\textbf{33.04}}\\
  & & &\cc QKV4 &\cc 39.65 &\cc 27.70 &\cc 22.60 &\cc 26.02 &\cc 64.56 &\cc 5.47 &\cc \textcolor{red}{\textbf{31.83}}\\
  \midrule
\multirow{20}{*}{\textbf{openPangu-7B}} & - &  BF16 & - & 36.25 & 24.78 & 20.50 & 41.84 & 60.70 & 23.96 & 34.45 \\ \cline{2-11}
  & \multirow{9}{*}{W8A8} & \multirow{3}{*}{INT8} & - & 37.22 & 26.20 & 20.29 & 40.94 & 60.41 & 26.85 & \textcolor{blue}{\textit{34.93}} \\
  & & & Q16KV8 & 36.83 & 26.71 & 20.44 & 41.30 & 59.72 & 25.70 & \textcolor{blue}{\textit{34.84}} \\
  & & & QKV8 & 36.86 & 26.94 & 20.49 & 40.05 & 61.81 & 26.60 & \textcolor{red}{\textbf{35.05}}  \\ \cline{3-11}
  & & \multirow{3}{*}{MXFP8} & - & 36.73 & 26.21 & 20.45 & 42.54 & 60.00 & 27.15 & \textcolor{red}{\textbf{35.14}} \\
  & & & Q16KV8 & 37.12 & 26.54 & 20.55 & 41.10 & 57.49 & 30.29 & \textcolor{red}{\textbf{34.97}}\\
  & & & QKV8 & 36.06 & 27.10 & 20.41 & 39.88 & 55.71 & 28.47 & \textcolor{blue}{\textit{34.18}} \\ \cline{3-11}
  & & \multirow{3}{*}{HiF8} &\cc - &\cc 36.03 &\cc 25.49 &\cc 20.37 &\cc 42.67 &\cc 59.72 &\cc 21.57 &\cc 34.31\\
  & & &\cc Q16KV8 &\cc 36.27 &\cc 25.34 &\cc 20.63 &\cc 38.53 &\cc 53.51 &\cc 23.27 &\cc 32.86 \\
  & & &\cc QKV8 &\cc 35.42 &\cc 25.40 &\cc 20.47 &\cc 39.00 &\cc 55.65 &\cc 23.92 &\cc 33.14 \\
  \cline{2-11}
  & \multirow{12}{*}{W4A4} & \multirow{3}{*}{INT4} & - & 14.09 & 8.51 & 14.60 & 5.34 & 5.03 & 10.26 & 9.80 \\
  & & & Q16KV4 & 3.69 & 3.16 & 12.51 & 2.09 & 2.20 & 13.54 & 5.69\\
  & & & QKV4 & 3.56 & 3.28 & 12.71 & 2.13 & 2.88 & 13.54 & 5.83 \\ \cline{3-11}
  & & \multirow{3}{*}{MXFP4} & - & 33.14 & 20.18 & 20.51 & 29.81 & 58.86 & 27.63 & 30.78 \\
  & & & Q16KV4 & 26.00 & 14.91 & 19.64 & 12.21 & 23.75 & 18.59 & 19.02 \\
  & & & QKV4 & 21.89 & 13.35 & 19.23 & 11.33 & 17.85 & 15.79 & 16.59\\ \cline{3-11}
  & & \multirow{3}{*}{NVFP4} & - & 35.35 & 23.78 & 20.53 & 37.71 & 59.71 & 16.04 & \textcolor{blue}{\textit{32.41}} \\
  & & & Q16KV4 & 32.97 & 23.14 & 20.11 & 26.34 & 55.21 & 15.13 & \textcolor{blue}{\textit{28.86}} \\
  & & & QKV4 & 30.08 & 22.07 & 20.00 & 25.42 & 51.27 & 14.31 & \textcolor{blue}{\textit{27.27}}\\ \cline{3-11}
  & & \multirow{3}{*}{HiF4} &\cc - &\cc 36.28 &\cc 25.81 &\cc 20.40 &\cc 39.34 &\cc 61.09 &\cc 23.34 &\cc \textcolor{red}{\textbf{34.16}}\\
  & & &\cc Q16KV4 &\cc 35.14 &\cc 22.96 &\cc 20.24 &\cc 32.48 &\cc 54.76 &\cc 30.69 &\cc \textcolor{red}{\textbf{31.85}} \\
  & & &\cc QKV4 &\cc 34.74 &\cc 25.63 &\cc 20.18 &\cc 33.34 &\cc 53.01 &\cc 26.46 &\cc \textcolor{red}{\textbf{31.79}} \\
\bottomrule
\end{tabular}
}
\caption{LongBench Performance Evaluation of KV Cache and Attention Quantization with Various Formats based on the Weight and Activation Quantization using SmoothQuant. Q16KV8 and Q16KV4 denote KV cache quantization, while QKV8 and QKV4 simulate attention quantization. Accuracy Loss denotes the average accuracy loss compared with the BF16 baselines. \textcolor{red}{\textbf{Bold red}} and \textcolor{blue}{\textit{italic blue}} indicate the best and second best settings, respectively.}
\label{tab:results-main_qkv_attention_quant_longbench}
\end{table*}

The activation analysis in Figure~\ref{fig:dist_4bit_a} demonstrates that managing dynamic outliers is the primary challenge for 4-bit inference. NVFP4 achieves the most faithful reconstruction of the original signal, due to its FP32 global scaling factor and fine-grained 16-element block size. This localized scaling effectively isolates high-magnitude outliers, preventing them from diluting the precision of the remaining values within the same block. HiF4 remains highly competitive but exhibits a slight loss of precision near zero; its coarser binning in the central region causes a high density of small-magnitude values to collapse into a limited number of states. This sensitivity is exacerbated by its larger block size relative to NVFP4, where a single extreme outlier can exert a disproportionate influence on the scaling factor for the entire group. To further investigate the impact of outliers on block-based quantization formats, we visualize a 64-element segment of the activation vector from the Qwen3-8B Layer 3 \texttt{down\_proj} along with its corresponding quantization results in Figure~\ref{fig:block_density}. The magnified view of indices 16--31 demonstrates that NVFP4, bolstered by its FP32 global scale, successfully avoids the range inflation that typically plagues INT4. Although HiF4 lacks a global scaling factor, its three-level hierarchy effectively isolates local variance, thereby preserving a higher number of active quantization bins compared to MXFP4.

The distribution of Key states in Figure~\ref{fig:dist_4bit_k} is characterized by a broad range and a spiky, multi-modal structure. This sparsity makes it the primary bottleneck for KV quantization. NVFP4 demonstrates the most precise alignment with this signal, as its 16-element block size allows the quantization grid to adapt rapidly to local magnitude shifts between different spikes. HiF4 remains highly resilient but exhibits a slight loss of resolution in the low-magnitude valleys between peaks. Because its hierarchical scaling must occasionally account for multiple peaks within a single block, the resulting grid can become slightly less dense in the regions between them. In contrast, MXFP4 and INT4 struggle significantly; their coarser scaling mechanisms are unable to track these rapid fluctuations, leading to a blurring of the Key distribution that likely contributes to the performance degradation seen in long-context benchmarks.

In contrast to the Key states, the Value distributions in Figure~\ref{fig:dist_4bit_v} appear more bounded and stable, mirroring the bell-shaped density often found in weights. In this more predictable numerical space, the performance gap between formats begins to narrow, as the primary requirement is representational density rather than expansive dynamic range. HiF4 and NVFP4 both maintain a high concentration of representable states within the primary bulk of the Value distribution, ensuring that the majority of the signal is captured with high fidelity. However, INT4 still experiences a significant resolution loss, as 16 levels are insufficient to represent the nuanced variations within even a bounded Value tensor. MXFP4 also shows wider intervals than the hierarchical formats, confirming that even for well-behaved distributions like Values, a multi-level scaling approach is superior to a single-level power-of-two scale.

\subsection{Additional Results}\label{sec:additional}
Table\ref{tab:results-main_qkv_attention_quant} provides a comprehensive performance evaluation of KV cache and attention quantization for the Qwen3-8B and openPangu-7B models. In the W8A8 regime, both models demonstrate high stability across all formats, with MXFP8 and HiF8 maintaining near-lossless performance. The transition to W4A4 precision coupled with KV cache and attention quantization reveals a critical bifurcation in format resilience. Traditional INT4 quantization suffers a catastrophic failure. While MXFP4 remains functional under weight and activation quantization alone, the addition of 4-bit KV cache and attention quantization triggers a severe performance decline. In contrast, HiF4 exhibits remarkable robustness under these extreme constraints. For Qwen3-8B, HiF4 consistently yields the superior result, restricting accuracy loss to 3.15\% for Q16KV4 and 3.92\% for QKV4. This trend is mirrored in the openPangu-7B results, where HiF4 maintains significantly higher model integrity than both MXFP4 and NVFP4 by limiting the loss to 11.03\% and 13.84\% respectively. These findings validate HiF4 as a uniquely effective format for end-to-end 4-bit inference, ensuring functional stability even when the entire model pipeline is subjected to ultra-low bit-width quantization. As shown in Table~\ref{tab:results-main_qkv_attention_quant_longbench}, the results from the LongBench dataset corroborate the findings from standard benchmarks, confirming that the hierarchical stability of HiF4 is critical for maintaining performance over long-context tasks, where the error accumulation may amplify the quantization errors caused by different low-bit data formats. HiF4 noticeably outperforms NVFP4 in term of the accuracy under the full 4-bit WA and KV cache quantization for both Qwen3-8B and openPanu-7B models.

\end{document}